\documentclass{article} 
\usepackage{iclr2027_conference,times}

\usepackage{amsmath,amsfonts,bm}

\def\eqref#1{equation~\ref{#1}}

\def\1{\bm{1}}

\DeclareMathAlphabet{\mathsfit}{\encodingdefault}{\sfdefault}{m}{sl}
\SetMathAlphabet{\mathsfit}{bold}{\encodingdefault}{\sfdefault}{bx}{n}

\usepackage{amssymb}
\usepackage{graphicx}
\usepackage{float}
\usepackage{xcolor}
\usepackage{colortbl}
\usepackage{hyperref}
\usepackage{url}
\usepackage[most]{tcolorbox}
\usepackage[utf8]{inputenc}
\usepackage{newunicodechar}
\newunicodechar{，}{,}

\definecolor{my_red}{rgb}{0.835, 0.309, 0.243}
\definecolor{my_green}{rgb}{0.333, 0.635, 0.345}
\definecolor{my_blue}{rgb}{0.298, 0.505, 0.913}
\definecolor{my_yellow}{rgb}{0.945, 0.741, 0.278}
\definecolor{my_purple}{rgb}{0.356, 0.078, 0.431}

\newtcolorbox{promptbox}[1]{%
  enhanced,
  unbreakable,
  width=0.94\linewidth,
  center,
  colback=black!2,
  colframe=my_blue,
  colbacktitle=my_blue!8,
  coltitle=black,
  fonttitle=\bfseries,
  fontupper=\small,
  title={#1},
  boxrule=0.7pt,
  arc=1.5mm,
  left=6pt, right=6pt, top=5pt, bottom=5pt,
  before skip=8pt, after skip=10pt
}

\title{Spatial Memory Intelligence: Endowing World Models with Understanding-Driven Long-Term Memory}

\author{%
\textbf{Ying Yang}$^{1,*}$ \quad \textbf{Guiyu Zhang}$^{1,2,*}$ \quad \textbf{Lianghua Huang}$^{2}$ \quad \textbf{Chang Nie}$^{4}$ \\[3pt]
\textbf{Chenyang Si}$^{4}$ \quad \textbf{Haofan Wang}$^{5}$ \quad \textbf{Shaoshuai Shi}$^{6}$ \quad \textbf{Li Jiang}$^{1,3,\dagger}$
}

\makeatletter
\renewcommand{\@maketitle}{%
  \begingroup
  \centering
  \setlength{\parskip}{0pt}%
  \hrule height 3pt
  \vskip 10pt
  {\fontsize{16.5}{19.5}\selectfont\bfseries
   Spatial Memory Intelligence: Endowing World Models\\
   with Understanding-Driven Long-Term Memory\par}
  \vskip 10pt
  \hrule height 0.8pt
  \vskip 12pt
  {\fontsize{10}{13}\selectfont\@author\par}
  \vskip 6pt
  {\fontsize{9}{11.5}\selectfont
   $^{1}$The Chinese University of Hong Kong, Shenzhen\par
   $^{2}$Alibaba Group \quad $^{3}$Shenzhen Loop Area Institute \quad $^{4}$Nanjing University\par
   $^{5}$Lovart AI \quad $^{6}$Voyager Research, Didi Chuxing\par}
  \vskip 9pt
  {\hypersetup{hidelinks}\normalsize\textbf{Project page:}~\href{https://spatial-memory-intelligence.github.io/\#top}{\textcolor{my_green}{https://spatial-memory-intelligence.github.io/}}\par}
  \vskip 8pt
  \endgroup
}
\makeatother

\iclrfinalcopy
\begin{document}

\maketitle
\pagestyle{plain}
\thispagestyle{plain}
\begingroup
\renewcommand{\thefootnote}{\fnsymbol{footnote}}
\footnotetext[0]{\textsuperscript{*}Equal contribution.\quad\textsuperscript{$\dagger$}Corresponding author.}
\endgroup

\begin{abstract}
  \begingroup
  \renewcommand{\section}[1]{}%
  \section{Abstract}
Long-video generation and world models have shown strong potential for interactive entertainment and embodied simulation by predicting future observations conditioned on user actions and historical memory. However, as memory sequences grow longer and their structures become increasingly complex, managing long-range spatial context becomes increasingly challenging, calling for a more intelligent and systematic memory-management strategy. Building on the advancing spatial reasoning capabilities of multimodal large language models (MLLMs) and the broader vision of unified models, we propose Spatial Memory Intelligence (SMI), the first framework to systematically employ an understanding model for spatial-memory management in long-video world models. SMI introduces four coordinated atomic operations: spatial clustering, within-cluster sparsification, action-aware retrieval, and reliability-aware filtering.  Extensive experiments across multiple baselines, benchmarks, and world-model backbones demonstrate the effectiveness and generalizability of SMI, achieving comprehensive improvements in memory sparsity, generation stability, and spatial consistency.

  \endgroup

\end{abstract}

\begin{figure}[H]
\centering
\includegraphics[width=0.97\textwidth]{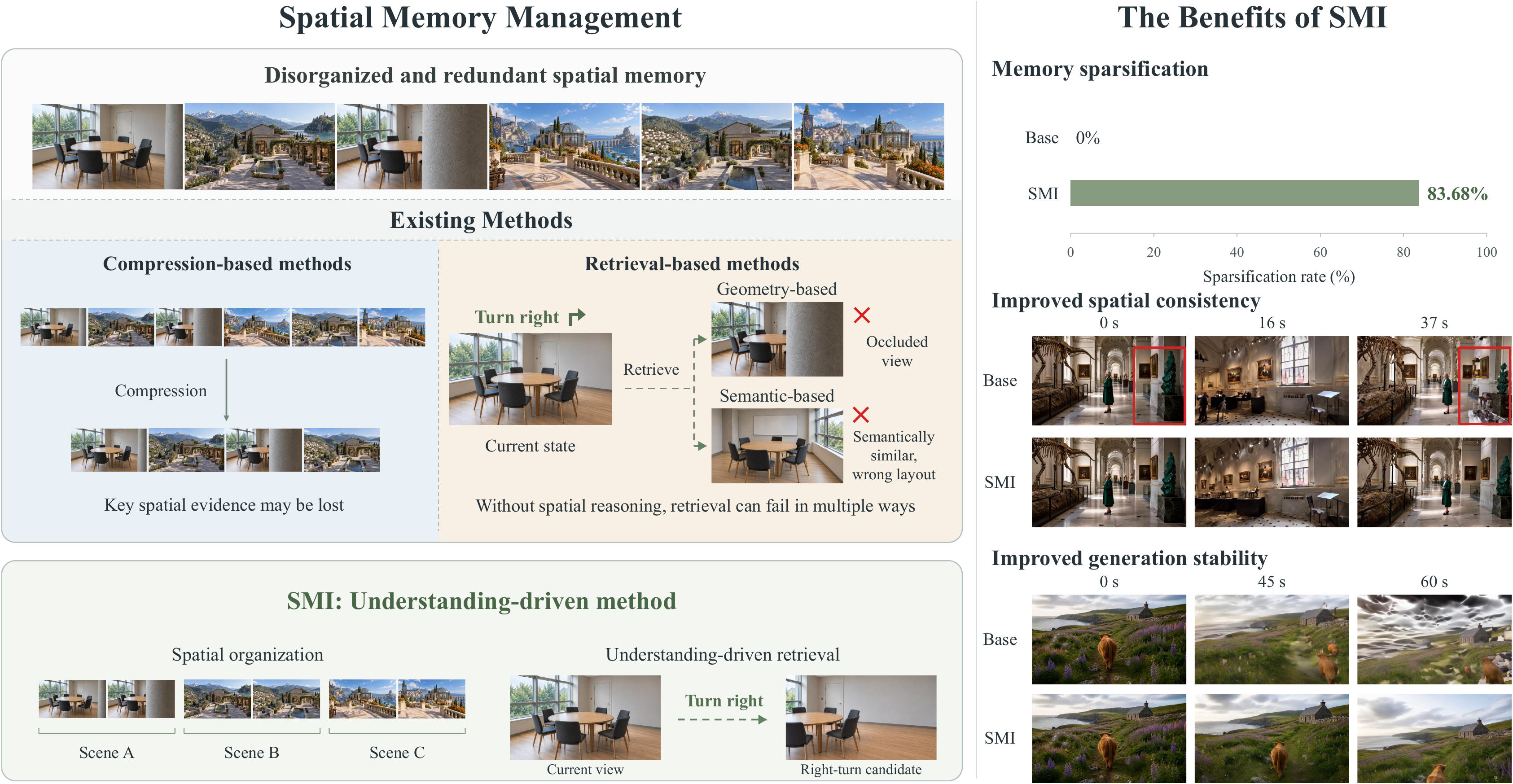}
\caption{\textbf{Motivation and benefits of SMI.} Understanding-driven spatial-memory management organizes disorganized and redundant histories and retrieves relevant memories, enabling 83.68\% memory sparsification with improved spatial consistency and generation stability.}
\label{fig:smi_teaser}
\end{figure}

\section{Introduction}

World models are driving a pivotal shift in visual and spatial intelligence, moving generative modeling toward controllable simulation of environments. Within this paradigm, video provides a natural medium for representing both the visual appearance and temporal evolution of an environment. Video world models~\citep{ball2025genie,che2025gamegen,sun2025worldplay,wang2026matrix,mao2026yume1,gao2026infinite} serve as interactive visual simulators by conditioning future visual states on historical observations, language instructions, and camera signals.
This interactivity enables spatial exploration across different regions of the generated world, where newly generated observations are continuously appended to the visual history and become part of the context for future prediction.
Through this iterative process, the model progressively integrates observations from different regions and viewpoints into an evolving world state that captures semantics and geometry, making video world models a promising foundation for interactive entertainment~\citep{tang2025hunyuan,team2026dreamx} and embodied simulation~\citep{agarwal2026cosmos,chen2026abot}.

The continual evolution of the generated world poses intertwined challenges for memory efficiency, spatial consistency, and generation stability. 
As the observation history grows, retaining all observations introduces substantial redundancy and incurs high storage costs, motivating memory sparsification~\citep{yu2025videossm,zhang2026frame}. In parallel, maintaining spatial consistency requires retrieving memories of previously observed regions that provide relevant semantic and geometric evidence. Failure to retrieve such evidence can lead to geometric inconsistencies in revisited regions and unintended changes in object identities or states~\citep{king2026echo,wu2026infinite,hu2026multiplayer}. 
Moreover, newly generated observations may themselves contain errors. If admitted into memory and reused as references, these errors can propagate and compound during autoregressive generation, undermining long-horizon stability~\citep{liu2025rolling,zhang2026frame}.
Together, these challenges call for systematic spatial-memory management that determines what to retain, what to retrieve, and which new observations are reliable enough to store.

Existing approaches to memory management in world models primarily focus on memory compression, sparse computation, and selective retrieval.
Compression-based methods~\citep{jin2025pyramidal,savov2026statespacediffuser,wu2026infinite,hong2025relic,zhang2026tinyhistory,wei2026geometry} summarize an expanding context into a bounded representation, often discarding 
fine-grained visual evidence needed to maintain long-horizon consistency.
Sparse-computation methods~\citep{xi2025sparse,xia2025training} exploit attention sparsity to reduce computational cost, but target generation efficiency rather than long-term memory management.
To maintain spatial consistency, retrieval-based methods select relevant historical observations as context based on geometric relations or semantic similarity.
Geometry-based retrieval~\citep{yu2025context,xiao2026worldmem,wu2026video,ren2025gen3c,li2025vmem} relies on camera or geometry relevance and is sensitive to occlusion, dynamic content, semantic changes and geometry errors.
Semantic retrieval~\citep{ji2025memflow,wu2025pack} captures high-level relevance but can struggle to distinguish similar content across different locations or viewpoints. 
Some methods jointly address memory compression and implicit content querying~\citep{zhang2026tinyhistory} through an auxiliary encoder and the generative model itself, but remain constrained by their limited spatial perception and semantic understanding. 
These limitations motivate a unified memory manager that organizes, sparsifies, and retrieves spatial context while reasoning over both geometry and semantics.

\begin{figure}[t]
    \centering
    \includegraphics[width=\linewidth]{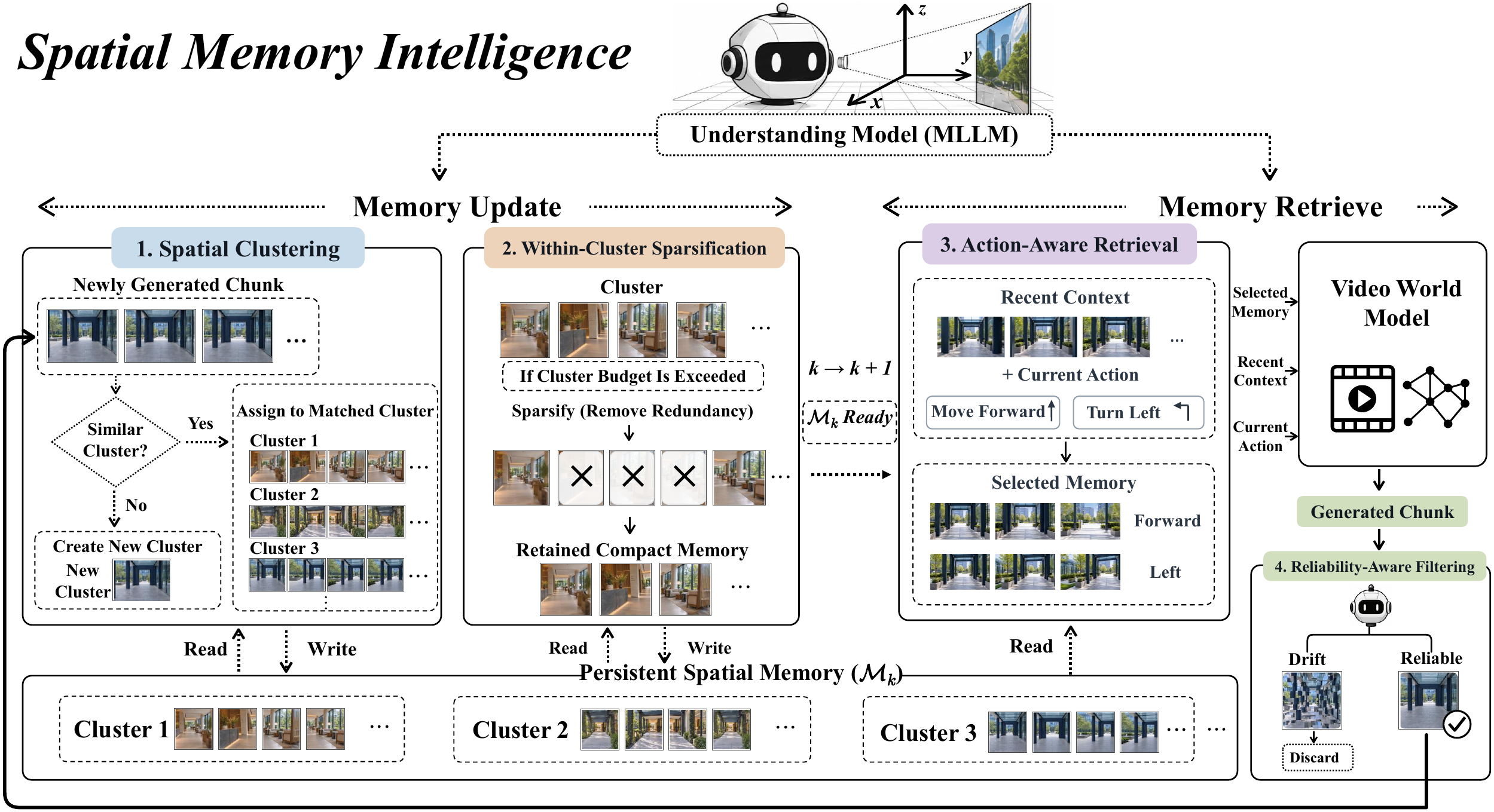}
    \caption{Overview of SMI. SMI uses an MLLM to manage spatial memory $\mathcal{M}_k$ through four atomic operations: spatial clustering, within-cluster sparsification, action-aware retrieval, and reliability-aware filtering.}
    \label{fig:framework}
\end{figure}

Recent MLLMs have made substantial progress in long-video understanding, spatial perception, and reasoning~\citep{bai2025qwen25vl,chen2024spatialvlm,yang2025thinkingspace,yang2025visual}.
Meanwhile, a growing line of work explores unified multimodal architectures that support both understanding and generation~\citep{wu2024janus,xie2024showo,deng2025bagel}.
\textbf{Motivated by the spatial and semantic capabilities of understanding models and by the broader vision of unified understanding and generation}, we propose Spatial Memory Intelligence (SMI), the first framework to systematically employ an understanding model for spatial-memory management in long-video world models.
SMI decomposes spatial-memory management into four coordinated atomic operations (Figure~\ref{fig:framework}): spatial clustering, within-cluster sparsification, action-aware retrieval, and reliability-aware filtering.
{Spatial clustering} structures the memory by grouping spatially proximate memory chunks into the same cluster. 
{Within-cluster sparsification} removes redundant observations from each cluster to reduce storage costs. 
{Action-aware retrieval} combines the recent context with the current action to determine which historical memories provide the most useful spatial evidence for generating the next chunk.  
{Reliability-aware filtering} prevents newly generated chunks exhibiting visual drift from entering memory, limiting error propagation through subsequent reuse.
Together, these atomic operations constitute a complete and unified mechanism for spatial-memory management.

For each operation, we formulate dedicated instructions and construct corresponding supervision, enabling the MLLM to learn the semantic and spatial judgments required throughout the memory-management pipeline. Compared with existing approaches, SMI provides a more comprehensive memory-management framework and achieves broader performance improvements, as illustrated in Figure~\ref{fig:smi_teaser}.
Our contributions are summarized as follows:
\vspace{-0.5em}
\begin{itemize}
\item We introduce SMI, the first unified MLLM-driven framework for systematic spatial-memory management in long-horizon video world models. SMI defines four coordinated atomic operations for organizing, consolidating, filtering, and recalling spatial memory.
\vspace{-0.2em}
\item We construct dedicated operation-oriented supervision datasets covering all four atomic operations and formulate these operations as learnable multimodal decision tasks, enabling the MLLM to develop unified reasoning capabilities for spatial-memory management.
\vspace{-0.2em}

\item Extensive experiments across multiple baselines, benchmarks, and world-model backbones demonstrate the effectiveness and general applicability of SMI, with consistent improvements in memory sparsity, long-horizon generation stability, and visual consistency.
\end{itemize}
\vspace{-0.2em}
\section{Related Work}
\label{sec:related}

\subsection{Video World Models}
Classical world models learn predictive environment representations for planning and control~\citep{ha2018world,hafner2023mastering}.
Enabled by scalable Diffusion Transformer architectures~\citep{peebles2023scalable}, video generation has advanced substantially~\citep{hacohen2024ltx,yang2025cogvideox,wan2025wan}, accelerating the development of video world models for interactive simulation of evolving environments.
Early video world models focused on controllable visual simulation in games and virtual environments~\citep{bruce2024genie,che2025gamegen,zhang2025matrix}.
Subsequent work has moved toward persistent world simulation, combining streaming generation~\citep{chen2024diffusion,huang2026self,yuan2026helios} with interactive control~\citep{yu2025gamefactorly,sun2025worldplay} to support long-horizon interaction in evolving environments~\citep{li2025hunyuan,hong2025relic,wang2026worldcompass,team2026advancing}.
Recent work has further extended video world modeling to multi-agent interaction~\citep{savva2026solaris,hu2026multiplayer,liu2026gamma} and embodied simulation for robotics and autonomous systems~\citep{pai2025mimic,agarwal2026cosmos,ye2026world,li2026causal}.
Despite these advances, persistent world simulation remains challenging, as long-horizon generation requires sparsifying an ever-growing observation history to reduce storage costs while preserving scene consistency and limiting error accumulation. 
This motivates spatial-memory management that jointly addresses memory sparsity, spatial consistency, and generation stability.

\subsection{Memory in Video Generation and World Models}

Long-horizon autoregressive generation accumulates observations across regions and viewpoints, increasing memory overhead and requiring generated content to remain spatially consistent when previously observed regions are revisited.
Existing methods for long-horizon context management can be broadly categorized into memory compression, sparse computation, and selective retrieval.
Compression-based methods~\citep{zhang2026frame,jin2025pyramidal,savov2026statespacediffuser,wu2026infinite,hong2025relic,zhang2026tinyhistory,wei2026geometry,zhu2025memorize} encode accumulated history into compact memory representations, reducing the amount of context retained over long horizons. However, compact representations may discard fine-grained visual and geometric information.
Sparse-computation methods~\citep{xi2025sparse,cai2023efficientvit,choromanski2020rethinking,xia2025training,cai2025mixture,glorian2026lvsa} exploit attention sparsity to reduce computational cost, focusing on efficient context processing rather than selective memory retention.
Retrieval-based methods select relevant historical observations according to geometric relations or semantic similarity.
Geometry-based retrieval~\citep{yu2025context,xiao2026worldmem,wu2026video,ren2025gen3c,li2025vmem,huang2025memoryforcing,joo2026retrieve} relies on camera or geometry relevance and is sensitive to occlusion, dynamic content, semantic changes, and geometry errors.
Semantic retrieval~\citep{ji2025memflow,wu2025pack} captures semantic relevance but can struggle to distinguish similar content across different locations or viewpoints.
Some methods~\citep{zhang2026tinyhistory} further combine memory compression with implicit retrieval through an auxiliary encoder and the generative model, but remain limited by their spatial perception and semantic understanding.
Overall, existing approaches address memory efficiency and retrieval through different mechanisms, but lack a unified framework for spatial-memory management.

\subsection{Multimodal intelligence}
Spatial-memory management inherently requires spatial perception, semantic understanding, and long-horizon reasoning. Recent advances in multimodal intelligence provide a natural foundation for these requirements. MLLMs have acquired increasingly strong capabilities in long-video reasoning~\citep{ren2024timechat,song2024moviechat,he2024mallm}. Meanwhile, VLMs have progressed from basic spatial perception toward reasoning over 3D and multi-view environments, including cognitive-map construction and continuous scene understanding~\citep{chen2024spatialvlm,ma2024spatialpin,ma2025spatialllm,yang2025thinkingspace,wang2025spatialclip,zhu2025cospace,yang2025visual}. Furthermore, recent models explore depth and multi-view inputs, geometry-aware representations, and spatially grounded learning strategies to improve spatial perception and reasoning~\citep{daxberger2025mmspatial,wu2025spatialmllm,ma2025spatialreasoner,song2025seetrek,shen2025spatialreasonerr1,zhao2026spacemind,liu2026spatialssrl}.
In parallel, unified multimodal models are beginning to dissolve the traditional boundary between understanding and generation~\citep{lu2024unifiedio2,jin2024videolavit,wu2024janus,xie2024showo,wu2025vilau,deng2025bagel}. These advances point toward a forward-looking paradigm in which understanding is not merely an auxiliary capability applied after generation, but an integral component that actively organizes and governs the generative process. 
\section{Method}
\label{sec:method}

\label{sec:our_work}

We propose Spatial Memory Intelligence (SMI), a unified framework that employs an MLLM to manage long-term spatial memory for video world models, as illustrated in Figure~\ref{fig:framework}. SMI comprises four coordinated atomic operations: spatial clustering, within-cluster sparsification, action-aware retrieval, and reliability-aware filtering. We begin by formalizing action-conditioned long-video generation in Section~\ref{sec:preliminaries} and explaining the motivation for using an MLLM as the spatial-memory manager in Section~\ref{sec:why_mllm}. We then introduce the four spatial-memory operations in Section~\ref{sec:atomic_ops}, followed by the operation-oriented data construction and training procedure in Section~\ref{sec:operation_training}.

\subsection{Preliminaries}
\label{sec:preliminaries}
\paragraph{Action-conditioned video world models.}
Action-conditioned video world models generate frames autoregressively from selected historical observations and current actions. Given an action sequence $a_{1:K}$, the joint distribution of the generated video frames $\boldsymbol{x}_{1:K}$ is formulated as
\begin{equation}
p_{\theta}(\boldsymbol{x}_{1:K}\mid a_{1:K})
=
\prod_{k=1}^{K}
p_{\theta}\left(
\boldsymbol{x}_k
\mid
\boldsymbol{x}_{<k}^{\mathrm{subset}},
a_k
\right),
\label{eq:action_conditioned_generation}
\end{equation}
where $\theta$ denotes the model parameters, $K$ is the number of generated frames, $\boldsymbol{x}_k$ is the frame generated at step $k$, $\boldsymbol{x}_{<k}^{\mathrm{subset}}$ denotes the selected historical frames before step $k$, and $a_k$ is the current action.
SMI manages the historical observations used in this process.

\begin{figure*}[t]
    \centering
    \includegraphics[width=\textwidth]{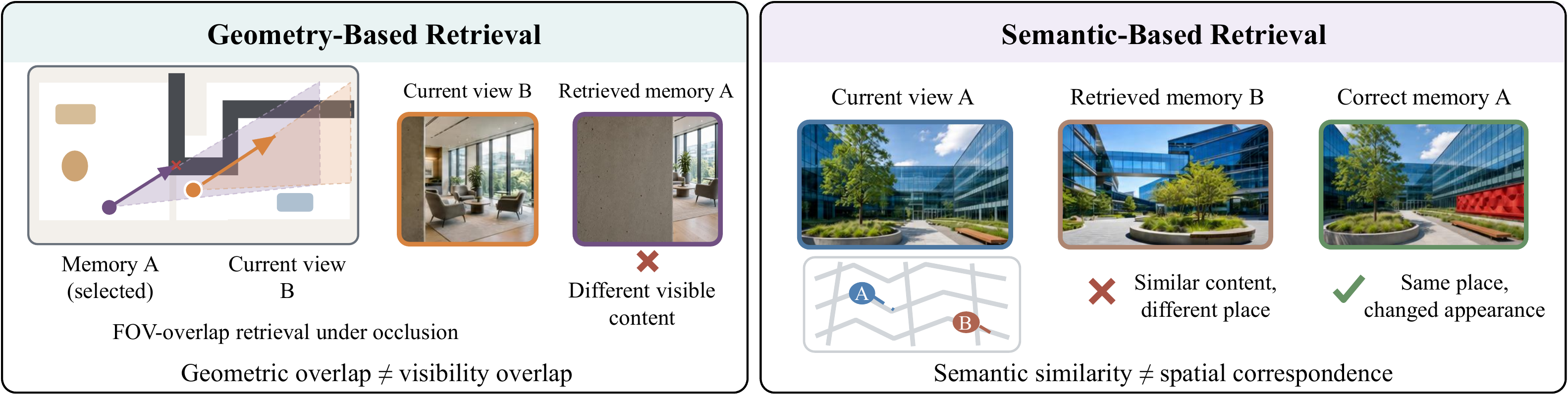}
    \caption{Failure modes of geometry- and semantic-based memory retrieval. }
    \label{fig:retrieval_failures}
\end{figure*}

\subsection{Why Use an MLLM?}
\label{sec:why_mllm}
Despite leveraging geometric or semantic information, existing memory-management strategies still exhibit retrieval limitations across diverse scenes, as illustrated in Figure~\ref{fig:retrieval_failures}.
Geometry-based retrieval can fail to identify useful memories under spatial occlusion, whereas semantic retrieval may overlook spatial and viewpoint changes.
The fundamental limitation is that these approaches lack the ability to understand the relationships between spatial observations and to reason based on these relationships.
Over long-horizon interaction, \textbf{managing an increasingly complex observation history requires reasoning over historical states, tracking semantic and spatial changes during interaction, and relating observations across viewpoints and time.}
Meeting these demands remains challenging for both fixed heuristics and the generator's implicit memory mechanisms.

These limitations call for a dedicated memory manager that can reason over the evolving history and determine how memory should be organized, retained, and retrieved for the current context.
Scalable, data-driven MLLMs provide a natural basis for such a manager, as they can learn memory-management strategies from diverse multimodal data.
Moreover, their growing capabilities in contextual understanding and semantic reasoning~\citep{alayrac2022flamingo,mckinzie2024mm1,bai2025qwen3}, together with rapid advances in spatial perception and reasoning~\citep{chen2024spatialvlm,yang2025thinkingspace,ma2025spatialllm}, make MLLMs well suited to spatial-memory management in video world models.
More broadly, progress toward unified understanding-and-generation architectures~\citep{wu2024janus,xie2024showo,deng2025bagel} reflects a growing trend toward closer integration of multimodal understanding and generation.
In this work, we establish an initial framework for managing spatial memory with an MLLM and define four atomic operations.

\subsection{Atomic Spatial Operations}
\label{sec:atomic_ops}

SMI decomposes spatial-memory management into four coordinated atomic operations: spatial clustering, within-cluster sparsification, action-aware retrieval, and reliability-aware filtering. At generation step $k$, we define $C_k$ as the newly generated video chunk and $\mathcal{M}_k=\{m_i\}_{i=1}^{N_k}$ as the spatial memory bank, where $m_i$ is the $i$-th retained video memory chunk and $N_k=|\mathcal{M}_k|$ is the number of retained chunks. When camera parameters are available, $\pi_i$ denotes the camera parameters associated with $m_i$ and serves as an auxiliary geometric cue.

\begin{figure*}[t]
    \centering
    \includegraphics[width=\textwidth]{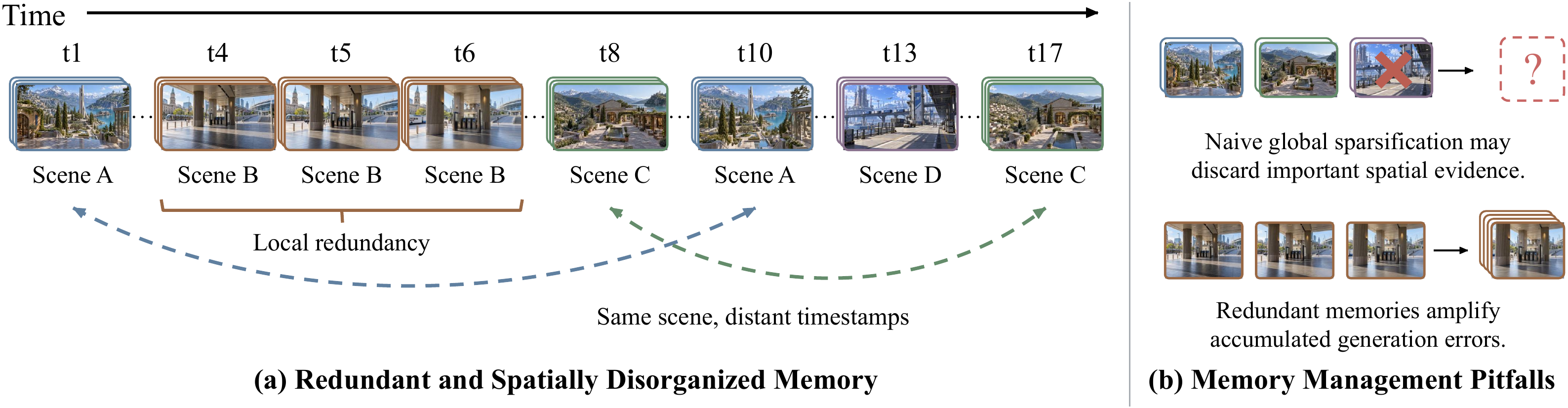}
    \caption{Motivation for spatial clustering: redundancy and spatial disorder in long-term memory.}
    \label{fig:clustering_motivation}
\end{figure*}

\paragraph{Spatial clustering.}
Temporal proximity does not imply spatial proximity. During inference, unpredictable user actions can cause the camera to revisit different regions in arbitrary order, interleaving spatially distinct observations in temporal memory. As shown in Figure~\ref{fig:clustering_motivation}, the resulting memory is temporally ordered but spatially disorganized and redundant, making sparsification difficult and causing generation errors to accumulate.
To address these issues, SMI organizes the streaming-generated chunks into spatial clusters. At step $k$, let the $j$-th spatial cluster be
\begin{equation}
\mathcal{S}_j=\{m_{j,q}\}_{q=1}^{n_j}\subseteq\mathcal{M}_k,
\label{eq:cluster_memory}
\end{equation}
where $j$ indexes the clusters, $q$ indexes the memory items in cluster $j$, and $n_j$ is the number of memory items in that cluster. We select the memory item with the largest projection coverage from the remaining chunks in the same cluster as its prototype:
\begin{equation}
p_j
=
\underset{m_{j,q}\in\mathcal{S}_j}{\operatorname{arg\,max}}
\operatorname{Cov}
\left(
m_{j,q},
\mathcal{S}_j-\{m_{j,q}\}
\right).
\label{eq:cluster_prototype}
\end{equation}
Here, $\mathcal{S}_j-\{m_{j,q}\}$ denotes all memory items in cluster $j$ except $m_{j,q}$, $\operatorname{Cov}$ denotes their average camera-projection coverage of $m_{j,q}$, and $p_j$ is the prototype of cluster $j$.

For a newly generated chunk $C_k$, we first use camera position and orientation to shortlist a set of spatially closest cluster prototypes. The MLLM then determines whether $C_k$ matches one of these candidate prototypes. If a match exists, $C_k$ is assigned to the corresponding spatial cluster; otherwise, a new cluster is created:
\begin{equation}
\operatorname{Assign}(C_k)
=
\begin{cases}
\mathcal{S}_j, & \text{if the MLLM matches } C_k \text{ to prototype } p_j,\\
\mathcal{S}_{\mathrm{new}}, & \text{if no matching prototype exists}.
\end{cases}
\label{eq:cluster_assignment}
\end{equation}
Here, $\mathcal{S}_{\mathrm{new}}$ denotes the new cluster created for $C_k$.

\paragraph{Within-cluster sparsification.}
After spatial clustering, the memory items within each cluster are ordered by generation time:
\begin{equation}
\mathcal{S}_j
=
\left(
m_{j,1},\ldots,m_{j,n_j}
\right).
\label{eq:ordered_cluster}
\end{equation}
When $n_j$ reaches a predefined sparsification trigger threshold $T_s$, SMI initiates within-cluster redundancy detection. The MLLM compares the intermediate memory items and identifies redundant chunks that can be safely removed. We formalize this operation as
\begin{equation}
\mathcal{K}_j
=
\operatorname{Sparse}_{\mathrm{MLLM}}
\left(
\mathcal{S}_j
\right),
\qquad
n_j\ge T_s.
\label{eq:sparse_decision}
\end{equation}
Here, $T_s$ specifies only the number of memory items required to trigger within-cluster sparsification, and $\mathcal{K}_j$ is the retained subset after redundant items are removed. The number of retained items is not fixed in advance; instead, it is determined dynamically by the MLLM according to whether each item contains new spatial structure, object states, or occlusion relations.  An intermediate item is removed if it provides no additional useful information relative to its neighboring memories. Because sparsification is performed only within spatially coherent clusters, SMI removes repeated observations without discarding important evidence from different spatial regions.

\paragraph{Action-aware retrieval.}
At generation step $k$, SMI selects useful historical memories conditioned on the recent context and current action. For efficiency, the current camera state and action are first used to shortlist the $K_g$ memory items with the largest spatial overlap with the current view:
\begin{equation}
\mathcal{G}_k
=
\operatorname{TopK}_{K_g}
\left(
\mathcal{M}_k;
\boldsymbol{\pi}_k,
a_k
\right).
\label{eq:geometric_candidates}
\end{equation}
Here, $k$ denotes the current generation step, $\boldsymbol{\pi}_k$ is the camera state at step $k$, $a_k$ is the current action, $K_g$ is the number of geometrically shortlisted candidates, and $\mathcal{G}_k$ is the resulting candidate set.

The MLLM then uses the recent context, current action, and candidate memories to determine which historical chunks provide useful information for generating the next chunk:
\begin{equation}
\mathcal{R}_k
=
\operatorname{Retrieve}_{\mathrm{MLLM}}
\left(
C_{k-1}^{\mathrm{recent}},
a_k,
\mathcal{G}_k
\right),
\qquad
0\le|\mathcal{R}_k|\le P_r.
\label{eq:retrieval_decision}
\end{equation}
Here, $C_{k-1}^{\mathrm{recent}}$ denotes the recent context, $\mathcal{R}_k$ is the retrieved historical memory set, and $P_r$ is the maximum number of returned memory items. The MLLM may return an empty set when none of the candidates is useful, or select up to $P_r$ memory items when multiple historical memories jointly support the next generation.

\paragraph{Reliability-Aware Filtering.}
For each newly generated video chunk \(C_k\), the MLLM determines whether it exhibits evident visual drift and outputs a filtering decision:
\begin{equation}
D_k
=
\operatorname{Filter}_{\mathrm{MLLM}}(C_k)
\in
\{\mathrm{admit},\mathrm{discard}\}.
\end{equation}
Here, \(\operatorname{Filter}_{\mathrm{MLLM}}\) denotes the reliability-aware filtering operation performed by the MLLM, and \(D_k\) denotes the resulting decision for \(C_k\). If \(D_k=\mathrm{admit}\), the chunk is added to $\mathcal{M}_k$; otherwise, it is discarded, thereby reducing the accumulation of generation errors.

We provide visualizations of the different atomic operations during actual inference in Appendix~\ref{app:qualitative_reviews}.

\subsection{Operation-Oriented Data Construction}
\label{sec:operation_training}

Although general-purpose MLLMs already possess a certain degree of spatial reasoning ability, these specialized spatial operations still need to be learned from data. We first collect approximately one thousand trajectories from real model inference and remove low-quality samples. Next, we manually annotate more than one hundred examples from the remaining samples. We then use these human-annotated examples as demonstrations for multiple state-of-the-art teacher MLLMs, which annotate the rest of the data. A detailed description of the data-construction pipeline and dataset composition is provided in Appendix~\ref{app:data_construction}.

Finally, human annotators review the labels generated by the teacher MLLMs, discard predictions with evident errors, and correct inconsistent annotations. Through this multi-stage data construction pipeline with repeated cleaning and filtering, we obtain a training dataset of $N_{\mathcal{D}}$ examples:
\begin{equation}
\mathcal{D}
=
\left\{
\left(\mathcal{I}_n,Y_n\right)
\right\}_{n=1}^{N_{\mathcal{D}}}.
\label{eq:operation_dataset}
\end{equation}
Here, $\mathcal{I}_n$ denotes the multimodal input for the corresponding atomic operation, and $Y_n$ denotes its annotated target output. We then optimize the MLLM using standard cross-entropy. For a target output sequence $Y_n$ with $L_n$ tokens, the training objective is
\begin{equation}
\mathcal{L}_{\mathrm{SFT}}(\phi)
=
-\frac{1}{N_{\mathcal{D}}}
\sum_{n=1}^{N_{\mathcal{D}}}
\sum_{\ell=1}^{L_n}
\log p_{\phi}\!\left(
y_{n,\ell}\mid \mathcal{I}_n,y_{n,<\ell}
\right).
\label{eq:sft_objective}
\end{equation}
Here, $\phi$ denotes the trainable parameters of the MLLM, $y_{n,\ell}$ denotes the $\ell$-th token in the target sequence of the $n$-th sample, and $y_{n,<\ell}$ denotes all target tokens preceding it. 

\section{Experiments}
\label{sec:experiments}


\subsection{Implementation Details}
\label{sec:implementation_details}
\paragraph{Backbones and optimization.}
We conduct experiments using the HY1.5-8B and Wan2.2-5B world-model backbones from WorldPlay-1.5~\citep{sun2025worldplay}.
Qwen3.5-4B~\citep{qwen3.5} serves as the multimodal understanding model for the proposed spatial-memory operations and is fine-tuned with AdamW using a learning rate of $2\times10^{-5}$ and a global batch size of 64.
Additional experimental details are provided in Appendix~\ref{app:experimental_details}.

\subsection{Evaluation Metrics}
\label{sec:evaluation_metrics}

We assess overall performance using four complementary evaluation protocols. 
\textbf{(1) VBench}~\citep{huang2024vbench}. We report background consistency, motion smoothness, and aesthetic quality to evaluate generation stability and consistency.
\textbf{(2) GPT-5.6-sol evaluation.} Three independent GPT-5.6-sol evaluator agents blindly assess the generated videos. We report Camera-Constrained Scene Consistency and Overall Visual Quality, with the detailed evaluation protocol and full results provided in Appendix~\ref{app:vlm_full}. 
\textbf{(3) Reconstruction consistency.} Following WorldPlay, each model traverses a round-trip camera trajectory, and we compute PSNR and LPIPS between corresponding frames from the outbound and return paths.
\textbf{(4) User study.} We conduct a user study to compare the generation stability and spatial consistency of SMI and the baselines. The evaluation protocol and results are provided in Appendix~\ref{app:user_study}. For VBench and GPT-5.6-sol evaluation, we use 100 randomly sampled one-minute rollouts. For reconstruction consistency, we use a separate set of 100 one-minute rollouts containing revisit trajectories. \textbf{For fair evaluation, all test scenes are held out from the training set. We will release all datasets and model weights for reproducibility.}

\begin{table*}[!t]
\begingroup
\centering
\caption{Comparison of the Base, five memory-management baselines, and SMI on HY1.5. All methods use identical generation settings. Bold indicates the best reported value.}
\label{tab:base_smi_results}
\scriptsize
\setlength{\tabcolsep}{2.8pt}
\resizebox{\textwidth}{!}{
\begin{tabular}{lccccccccc}
\hline
& \multicolumn{3}{c}{VBench}
& \multicolumn{2}{c}{GPT-5.6-sol evaluation}
& \multicolumn{2}{c}{Reconstruction}
& \multicolumn{2}{c}{Efficiency} \\
\cline{2-4}\cline{5-6}\cline{7-8}\cline{9-10}

\begin{tabular}[c]{@{}l@{}}Method\end{tabular}
& \begin{tabular}[c]{@{}c@{}}Background\\Consistency$\uparrow$\end{tabular}
& \begin{tabular}[c]{@{}c@{}}Motion\\Smoothness$\uparrow$\end{tabular}
& \begin{tabular}[c]{@{}c@{}}Aesthetic\\Quality$\uparrow$\end{tabular}
& \begin{tabular}[c]{@{}c@{}}Camera-Constrained\\Consistency$\uparrow$\end{tabular}
& \begin{tabular}[c]{@{}c@{}}Overall Visual\\Quality$\uparrow$\end{tabular}
& \begin{tabular}[c]{@{}c@{}}PSNR$\uparrow$\end{tabular}
& \begin{tabular}[c]{@{}c@{}}LPIPS$\downarrow$\end{tabular}
& \begin{tabular}[c]{@{}c@{}}Sparsity$\uparrow$\end{tabular}
& \begin{tabular}[c]{@{}c@{}}Latency$\downarrow$\end{tabular} \\
\hline

Base (FoV)~\citep{sun2025worldplay}
& 0.937983
& 0.991386
& 0.445389
& 51.6053
& 53.1538
& 12.478528
& 0.603425
& 0.0000\%
& $\times 1.00$ \\

FramePack~\citep{zhang2026frame}
& 0.939390
& 0.989362
& 0.432000
& 43.2800
& 46.6474
& 11.718369
& 0.621979
& \textbf{94.3820\%}
& $\times 1.00$ \\

Deep Forcing~\citep{yi2025deep}
& \textbf{0.946473}
& 0.982810
& 0.440462
& 37.6760
& 33.6297
& 11.260350
& 0.664867
& 79.2135\%
& $\mathbf{\times 0.81}$ \\

MoC~\citep{cai2025mixture}
& 0.937999
& 0.990528
& 0.438969
& 48.9196
& 51.9717
& 10.802151
& 0.735292
& 0.0000\%
& $\times 1.15$ \\

VMem~\citep{li2025vmem}
& 0.939695
& 0.991162
& 0.447630
& 54.4727
& 56.5060
& 12.768029
& 0.582628
& 0.0000\%
& $\times 1.47$ \\

MemFlow~\citep{ji2025memflow}
& 0.938170
& 0.990061
& 0.430962
& 43.3160
& 45.6500
& 10.296440
& 0.712713
& 49.4253\%
& $\times 1.27$ \\

\hline

\rowcolor{black!7}
SMI (ours)
& 0.942890
& \textbf{0.991999}
& \textbf{0.469357}
& \textbf{67.0200}
& \textbf{65.3677}
& \textbf{13.309393}
& \textbf{0.554519}
& 83.6842\%
& $\times 1.38$ \\

\hline
\end{tabular}
}
\endgroup


\begingroup
\centering
\caption{Comparison of the Base, five memory-management baselines, and SMI on Wan2.2. All methods use identical generation settings. Bold indicates the best reported value.}
\label{tab:main_results}
\scriptsize
\setlength{\tabcolsep}{2.8pt}
\resizebox{\textwidth}{!}{
\begin{tabular}{lccccccccc}
\hline
& \multicolumn{3}{c}{VBench}
& \multicolumn{2}{c}{GPT-5.6-sol evaluation}
& \multicolumn{2}{c}{Reconstruction}
& \multicolumn{2}{c}{Efficiency} \\
\cline{2-4}\cline{5-6}\cline{7-8}\cline{9-10}

\begin{tabular}[c]{@{}l@{}}Method\end{tabular}
& \begin{tabular}[c]{@{}c@{}}Background\\Consistency$\uparrow$\end{tabular}
& \begin{tabular}[c]{@{}c@{}}Motion\\Smoothness$\uparrow$\end{tabular}
& \begin{tabular}[c]{@{}c@{}}Aesthetic\\Quality$\uparrow$\end{tabular}
& \begin{tabular}[c]{@{}c@{}}Camera-Constrained\\Consistency$\uparrow$\end{tabular}
& \begin{tabular}[c]{@{}c@{}}Overall Visual\\Quality$\uparrow$\end{tabular}
& \begin{tabular}[c]{@{}c@{}}PSNR$\uparrow$\end{tabular}
& \begin{tabular}[c]{@{}c@{}}LPIPS$\downarrow$\end{tabular}
& \begin{tabular}[c]{@{}c@{}}Sparsity$\uparrow$\end{tabular}
& \begin{tabular}[c]{@{}c@{}}Latency$\downarrow$\end{tabular} \\
\hline

Base (FoV)~\citep{sun2025worldplay}
& 0.930115
& 0.928419
& 0.424737
& 34.6524
& 37.8363
& 11.434590
& 0.596041
& 0.0000\%
& $\times 1.00$ \\

FramePack~\citep{zhang2026frame}
& \textbf{0.934067}
& 0.933264
& 0.421781
& 34.7984
& 37.9917
& 11.011042
& 0.611023
& \textbf{94.3820\%}
& $\times 0.98$ \\

Deep Forcing~\citep{yi2025deep}
& 0.920607
& 0.929061
& 0.444197
& 46.8560
& 50.7300
& 11.198159
& 0.602744
& 79.2135\%
& $\mathbf{\times 0.75}$ \\

MoC~\citep{cai2025mixture}
& 0.916892
& 0.931185
& 0.432453
& 45.5013
& 49.2763
& 11.143642
& 0.610697
& 0.0000\%
& $\times 1.24$ \\

VMem~\citep{li2025vmem}
& 0.920369
& 0.929605
& 0.439904
& 43.8800
& 45.3613
& 11.157617
& 0.603184
& 0.0000\%
& $\times 1.34$ \\

MemFlow~\citep{ji2025memflow}
& 0.924393
& 0.932660
& 0.435056
& 38.8487
& 40.2990
& 10.887503
& 0.610859
& 49.4253\%
& $\times 1.13$ \\

\hline

\rowcolor{black!7}
SMI (ours)
& 0.917329
& \textbf{0.935492}
& \textbf{0.445095}
& \textbf{46.9309}
& \textbf{51.4909}
& \textbf{11.564875}
& \textbf{0.587550}
& 82.2073\%
& $\times 1.31$ \\

\hline
\end{tabular}
}
\endgroup
\end{table*}

\begin{figure*}[!t]
\begingroup
\centering
    \includegraphics[width=\textwidth]{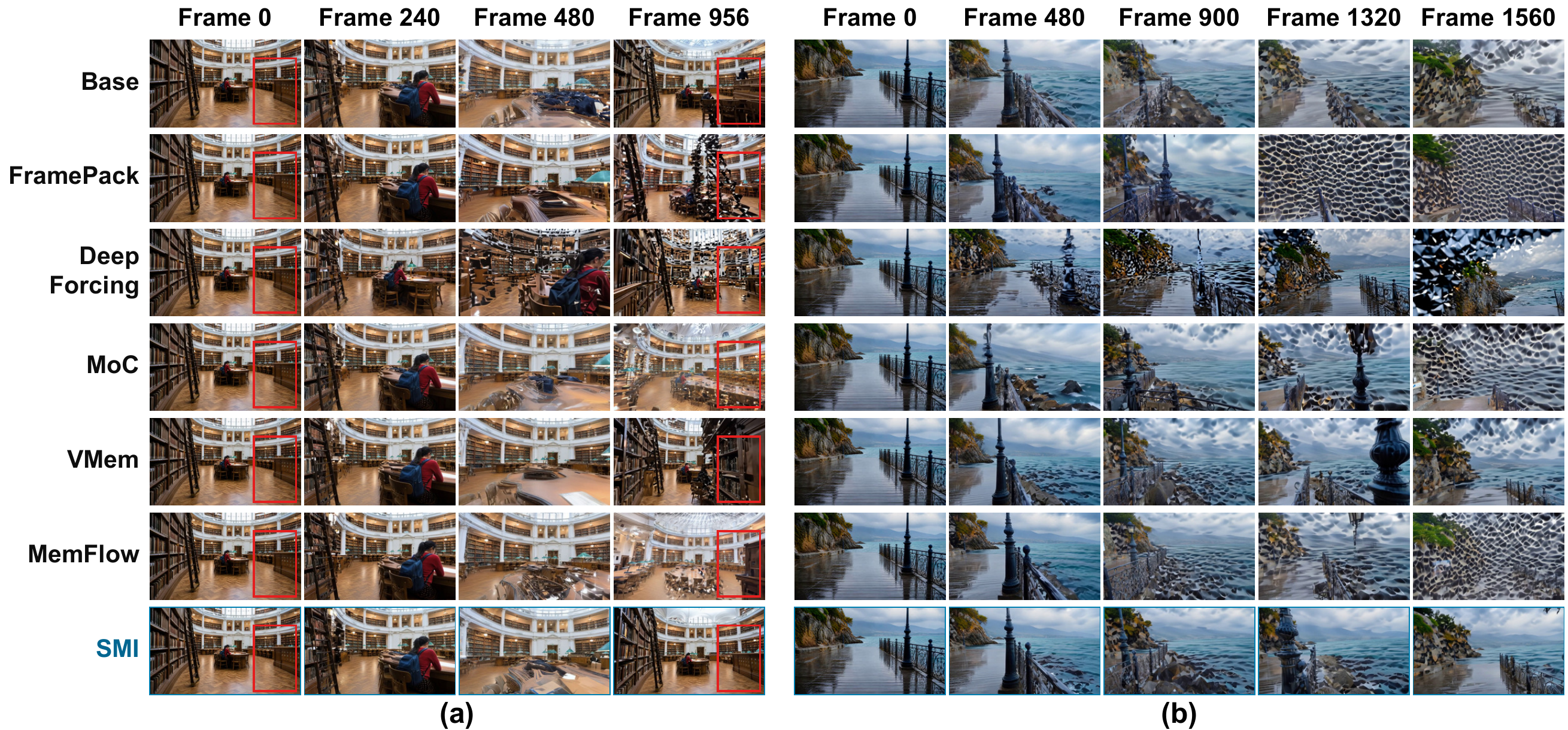}
    \caption{Qualitative comparisons on HY1.5: (a) spatial consistency and (b) generation stability. Red boxes highlight corresponding regions in the initial and final frames. }
    \label{fig:hy15_consistency_qualitative}
\endgroup

\vspace{2pt}

\begingroup
\centering
    \includegraphics[width=\textwidth]{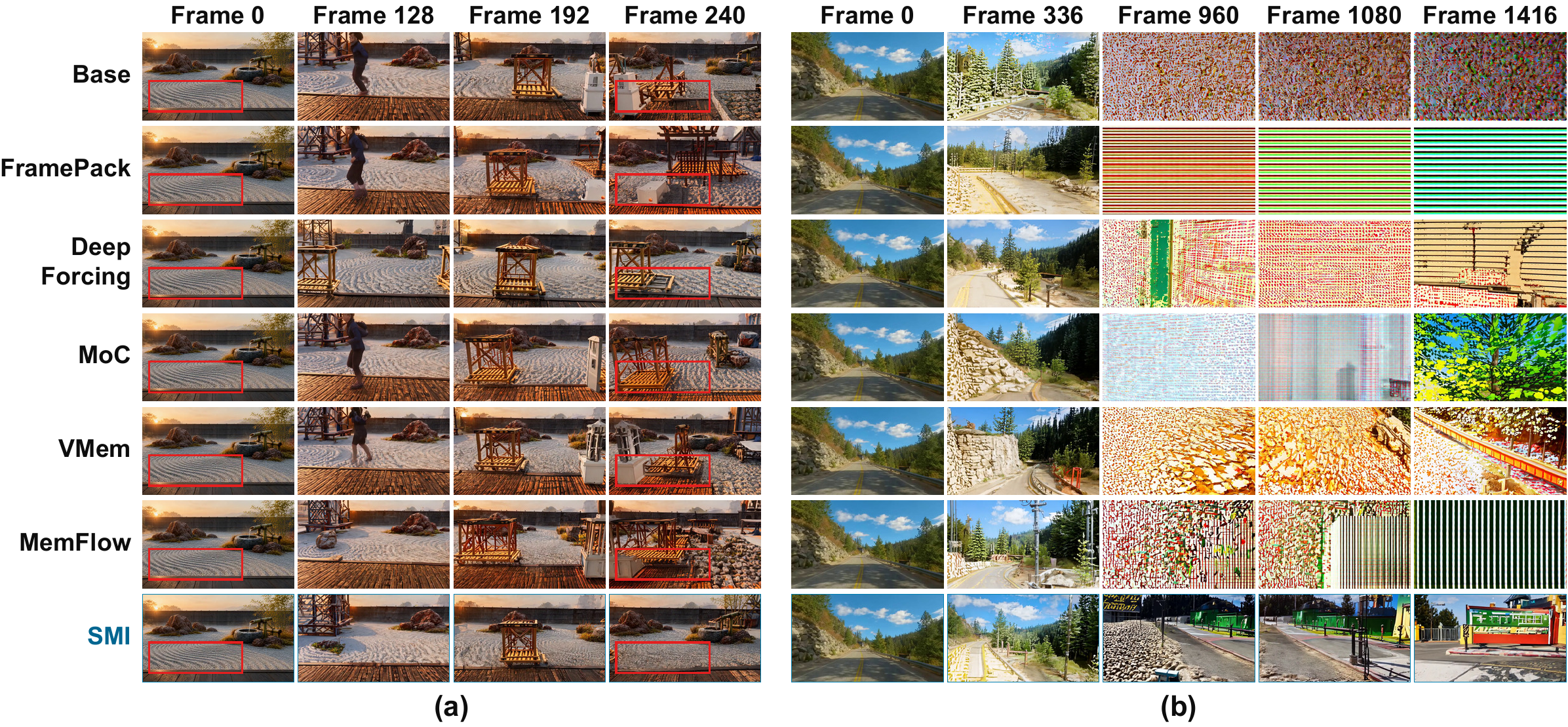}
    \caption{Qualitative comparisons on Wan2.2: (a) spatial consistency and (b) generation stability. Red boxes highlight corresponding regions in the initial and final frames. }
    \label{fig:wan22_seven_methods}
\endgroup
\end{figure*}
\vspace{-3pt}
\paragraph{Efficiency.}
We define memory sparsity as the proportion of memory items removed from the total memory. The Base performs no memory sparsification and therefore has 0\% sparsity. Latency is measured as end-to-end generation time relative to the Base, which is normalized to $\times 1.00$. All measurements use the same hardware and generation settings.
\vspace{-3pt}
\subsection{Baselines}
\label{sec:baselines}

Besides the HY1.5 and Wan2.2 baselines that use a field-of-view (FoV)-based retrieval strategy, we compare SMI with five memory-management methods: FramePack and Deep Forcing for compression, Mixture of Contexts (MoC) for sparse-routing-based retrieval, VMem for geometry-based retrieval, and MemFlow for semantic retrieval. These methods cover four complementary strategies for long-term memory management: compression, sparsification, geometric retrieval, and semantic retrieval. For a fair comparison, all methods use the same initial observations, action sequences, random seeds, context budget, and generation configuration. More details are provided in Appendix~\ref{app:baseline_details}.
\subsection{Results}
\label{sec:results}

\subsubsection{Quantitative Results}
Tables~\ref{tab:base_smi_results} and~\ref{tab:main_results} present quantitative comparisons between SMI and the baselines on HY1.5 and Wan2.2, respectively. On HY1.5, SMI improves Aesthetic Quality from 0.445389 to 0.469357, GPT-evaluated Scene Consistency from 51.6053 to 67.0200, and PSNR from 12.4785 to 13.3093 dB over the Base, while maintaining a high memory sparsity of 83.6842\%. A similar trend holds on Wan2.2. Overall, SMI substantially sparsifies memory while improving spatial consistency and generation stability, achieving more comprehensive improvements than the other methods. Although the additional understanding model introduces time overhead, we expect this overhead to decrease substantially as generation and understanding are unified through shared representations.
\subsubsection{Qualitative Results}
Figures~\ref{fig:hy15_consistency_qualitative} and~\ref{fig:wan22_seven_methods} illustrate the generation stability and spatial consistency of different methods on HY1.5 and Wan2.2. SMI demonstrates the best spatial consistency and generation stability among the compared methods on both backbones, achieving comprehensive improvements in both aspects. More qualitative results are provided in Appendix~\ref{app:additional_comparisons}.

\vspace{-5pt}
\subsection{Ablation Studies}
\label{sec:ablations}

\subsubsection{Component Ablation}
\begin{figure}[!htbp]
\centering
\includegraphics[width=\textwidth]{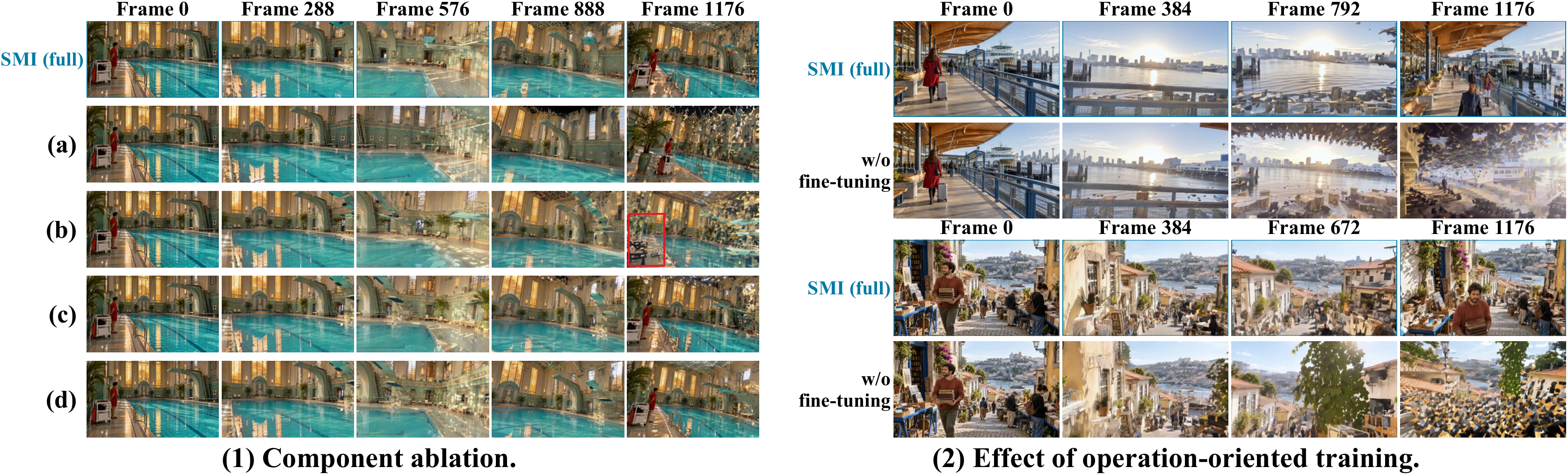}
\caption{\textbf{Qualitative ablations on HY1.5.} \textbf{(1) Component ablation:} SMI (full) is compared with variants without (a) reliability-aware filtering, (b) spatial clustering, (c)  spatial clustering and within-cluster sparsification, and (d) all three components. The red box highlights an inconsistent poolside region upon revisiting. \textbf{(2) Operation-oriented training:} SMI with the fine-tuned Qwen3.5-4B is compared with its counterpart without fine-tuning on two scenes.}
\label{fig:ablation_qualitative}
\end{figure}

We conduct ablations on HY1.5 by disabling different combinations of spatial clustering, within-cluster sparsification, and reliability-aware filtering, while retaining action-aware retrieval in all variants. Table~\ref{tab:component_ablation} and Figure~\ref{fig:ablation_qualitative}(1) present the quantitative and qualitative results. Removing components degrades generation stability and consistency to varying degrees. In particular, without spatial clustering, sparsification can discard important spatial evidence, leading to a decline in consistency. The reconstruction scores and the visual inconsistencies in Figure~\ref{fig:ablation_qualitative}(1) support this observation.


\begin{table*}[t]
\centering
\caption{Component ablation on HY1.5. A checkmark denotes an enabled module. Latency is relative to the Base. Bold indicates the best value.}
\label{tab:component_ablation}
\scriptsize
\setlength{\tabcolsep}{3pt}
\renewcommand{\arraystretch}{1.3}
\resizebox{\textwidth}{!}{
\begin{tabular}{ccccccccccc}
\hline
& \multicolumn{3}{c}{Modules}
& \multicolumn{3}{c}{VBench}
& \multicolumn{2}{c}{Reconstruction}
& \multicolumn{2}{c}{Efficiency} \\
\cline{2-4}\cline{5-7}\cline{8-9}\cline{10-11}

\begin{tabular}[c]{@{}c@{}}Setting\end{tabular}
& \begin{tabular}[c]{@{}c@{}}Spatial\\Clustering\end{tabular}
& \begin{tabular}[c]{@{}c@{}}Within-cluster\\Sparsification\end{tabular}
& \begin{tabular}[c]{@{}c@{}}Reliability-Aware\\Filtering\end{tabular}
& \begin{tabular}[c]{@{}c@{}}Background\\Consistency$\uparrow$\end{tabular}
& \begin{tabular}[c]{@{}c@{}}Motion\\Smoothness$\uparrow$\end{tabular}
& \begin{tabular}[c]{@{}c@{}}Aesthetic\\Quality$\uparrow$\end{tabular}
& \begin{tabular}[c]{@{}c@{}}PSNR$\uparrow$\end{tabular}
& \begin{tabular}[c]{@{}c@{}}LPIPS$\downarrow$\end{tabular}
& \begin{tabular}[c]{@{}c@{}}Sparsity$\uparrow$\end{tabular}
& \begin{tabular}[c]{@{}c@{}}Latency$\downarrow$\end{tabular} \\
\hline

\textbf{Full}
& $\checkmark$
& $\checkmark$
& $\checkmark$
& \textbf{0.942890}
& \textbf{0.991999}
& \textbf{0.469357}
& \textbf{13.309393}
& \textbf{0.554519}
& \textbf{83.6842\%}
& $\times 1.38$ \\
\hline

(a)
& $\checkmark$
& $\checkmark$
&
& 0.941184
& 0.991616
& 0.455922
& 13.232422
& 0.561628
& 81.5789\%
& $\times 1.36$ \\

(b)
&
& $\checkmark$
& $\checkmark$
& 0.939966
& 0.991568
& 0.450583
& 11.050455
& 0.675472
& 80.5263\%
& $\times 1.38$ \\

(c)
&
&
& $\checkmark$
& 0.939622
& 0.991600
& 0.457186
& 12.985179
& 0.567009
& 35.7368\%
& $\times 1.32$ \\

(d)
&
&
&
& 0.937700
& 0.991267
& 0.446470
& 12.812125
& 0.581249
& 0.0000\%
& $\mathbf{\times 1.27}$ \\
\hline
\end{tabular}
}
\end{table*}

\begin{table}[!htbp]
\centering
\caption{Effect of operation-oriented fine-tuning on HY1.5. Both variants use Qwen3.5-4B. Latency is relative to the Base; bold indicates the better value.}
\label{tab:ablation}
\label{tab:training_ablation}
\scriptsize
\setlength{\tabcolsep}{2.8pt}
\resizebox{\textwidth}{!}{
\begin{tabular}{lccccccccc}
\hline
& \multicolumn{3}{c}{VBench}
& \multicolumn{2}{c}{GPT-5.6-sol evaluation}
& \multicolumn{2}{c}{Reconstruction}
& \multicolumn{2}{c}{Efficiency} \\
\cline{2-4}\cline{5-6}\cline{7-8}\cline{9-10}

\begin{tabular}[c]{@{}l@{}}Variant\end{tabular}
& \begin{tabular}[c]{@{}c@{}}Background\\Consistency$\uparrow$\end{tabular}
& \begin{tabular}[c]{@{}c@{}}Motion\\Smoothness$\uparrow$\end{tabular}
& \begin{tabular}[c]{@{}c@{}}Aesthetic\\Quality$\uparrow$\end{tabular}
& \begin{tabular}[c]{@{}c@{}}Camera-Constrained\\Consistency$\uparrow$\end{tabular}
& \begin{tabular}[c]{@{}c@{}}Overall Visual\\Quality$\uparrow$\end{tabular}
& \begin{tabular}[c]{@{}c@{}}PSNR$\uparrow$\end{tabular}
& \begin{tabular}[c]{@{}c@{}}LPIPS$\downarrow$\end{tabular}
& \begin{tabular}[c]{@{}c@{}}Sparsity$\uparrow$\end{tabular}
& \begin{tabular}[c]{@{}c@{}}Latency$\downarrow$\end{tabular} \\
\hline

SMI (full)
& \textbf{0.942890}
& \textbf{0.991999}
& \textbf{0.469357}
& \textbf{67.0200}
& \textbf{65.3677}
& \textbf{13.309393}
& \textbf{0.554519}
& 83.6842\%
& $\mathbf{\times 1.38}$ \\

w/o fine-tuning
& 0.941856
& 0.991978
& 0.451266
& 42.8380
& 44.7440
& 9.396166
& 0.714904
& \textbf{86.3444\%}
& $\times 1.42$ \\
\hline
\end{tabular}
}
\end{table}

\subsubsection{Effect of Operation-Oriented Training}
We compare the original Qwen3.5-4B with its operation-oriented fine-tuned counterpart in SMI. As shown in Table~\ref{tab:training_ablation} and Figure~\ref{fig:ablation_qualitative}(2), the understanding model without fine-tuning struggles to reliably perform the atomic operations, resulting in substantially lower generation quality and consistency. Fine-tuning on high-quality data enables the understanding model to adapt to these operations and manage complex spatial memory.
\section{Conclusion}
\label{sec:conclusion}

In this work, we presented Spatial Memory Intelligence(SMI), the first unified framework that uses an understanding model to manage spatial memory in long-horizon video world models. SMI organizes memory through four coordinated atomic operations: spatial clustering, within-cluster sparsification, action-aware retrieval, and reliability-aware filtering. By sparsifying redundant memory observations while preserving useful spatial evidence, SMI substantially reduces memory overhead and improves long-horizon generation stability and spatial consistency. Extensive experiments across multiple baselines, benchmarks, and world-model backbones validate the effectiveness of SMI. Looking ahead, we hope this work takes a step toward a more general framework for spatial-memory management in future unified models.

\section{Limitations and Future Work}
\label{app:limitations_future_work}

\paragraph{Limitations.}
SMI has several limitations. First, the current framework is not trained end to end with the world model and therefore does not directly improve the intrinsic generation capability of the underlying backbone. Its performance remains bounded by the visual quality, controllability, and stability of the base generator. Second, because the understanding model and the generation model operate with different representations, the MLLM-based memory controller introduces additional computation, particularly during the early stages of a rollout before memory sparsification yields substantial savings. Third, the current task setting does not explicitly address long-horizon streaming understanding and may therefore remain insufficient for memory tasks involving complex long-term dependencies or logical reasoning.

\paragraph{Future work.}
Based on these limitations, several directions can be explored. First, generation, understanding, and memory management could be jointly learned within a shared representation space. Such a formulation may not only reduce computational overhead and enable end-to-end optimization, but also allow understanding and generation to mutually reinforce each other, thereby further enhancing the world model's modeling and memory capabilities. Second, reinforcement learning could be introduced into long-horizon world models with unified understanding and generation to further strengthen the understanding model's ability to manage long-term memory. Third, higher-quality and more comprehensive datasets could be constructed to enable understanding models to learn broader memory-management capabilities.

\bibliography{iclr2026_conference}
\bibliographystyle{iclr2027_conference}

\clearpage
\appendix
\section*{Appendix}

\section{Qualitative Examples of Runtime Memory Management}
\label{app:qualitative_reviews}

\begin{figure}[!ht]
    \centering
    \resizebox{!}{0.73\textheight}{%
        \vbox{\offinterlineskip
            \hbox{\includegraphics[trim=0bp 0bp 0bp 383.04bp,clip]{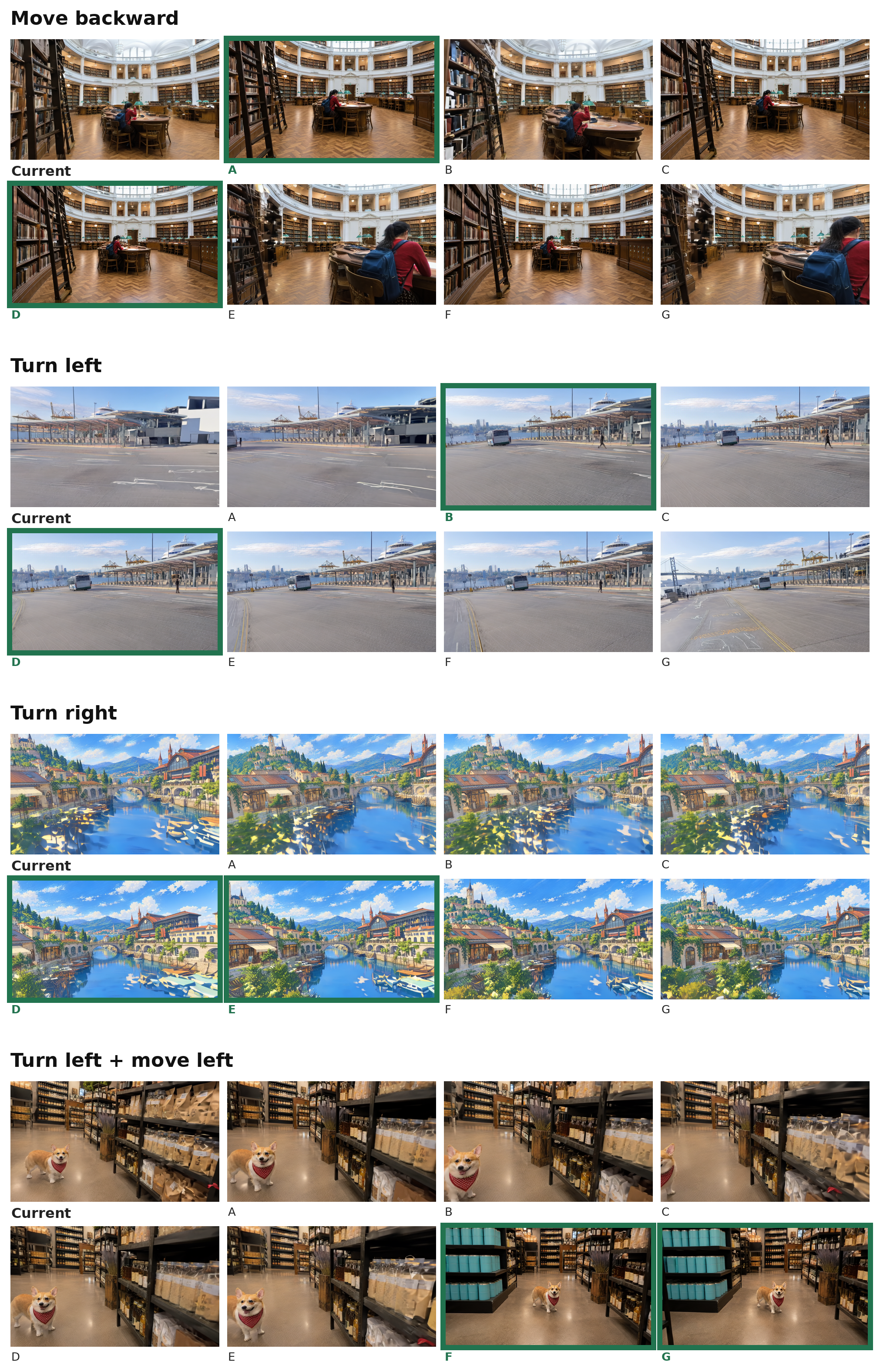}}%
            \vskip 9.6bp
            \hbox{\includegraphics[trim=0bp 383.04bp 0bp 0bp,clip]{sec/image/06_coverage_regroup_1_bold.png}}%
        }%
    }
    \caption{Action-aware retrieval examples. `Current' denotes the current observation, and A--G denote historical candidates. Green borders mark the selected candidates.}
    \label{fig:hy15_retrieval_runtime}
\end{figure}

\begin{figure}[!ht]
    \centering
    \includegraphics[width=\textwidth,height=0.92\textheight,keepaspectratio]{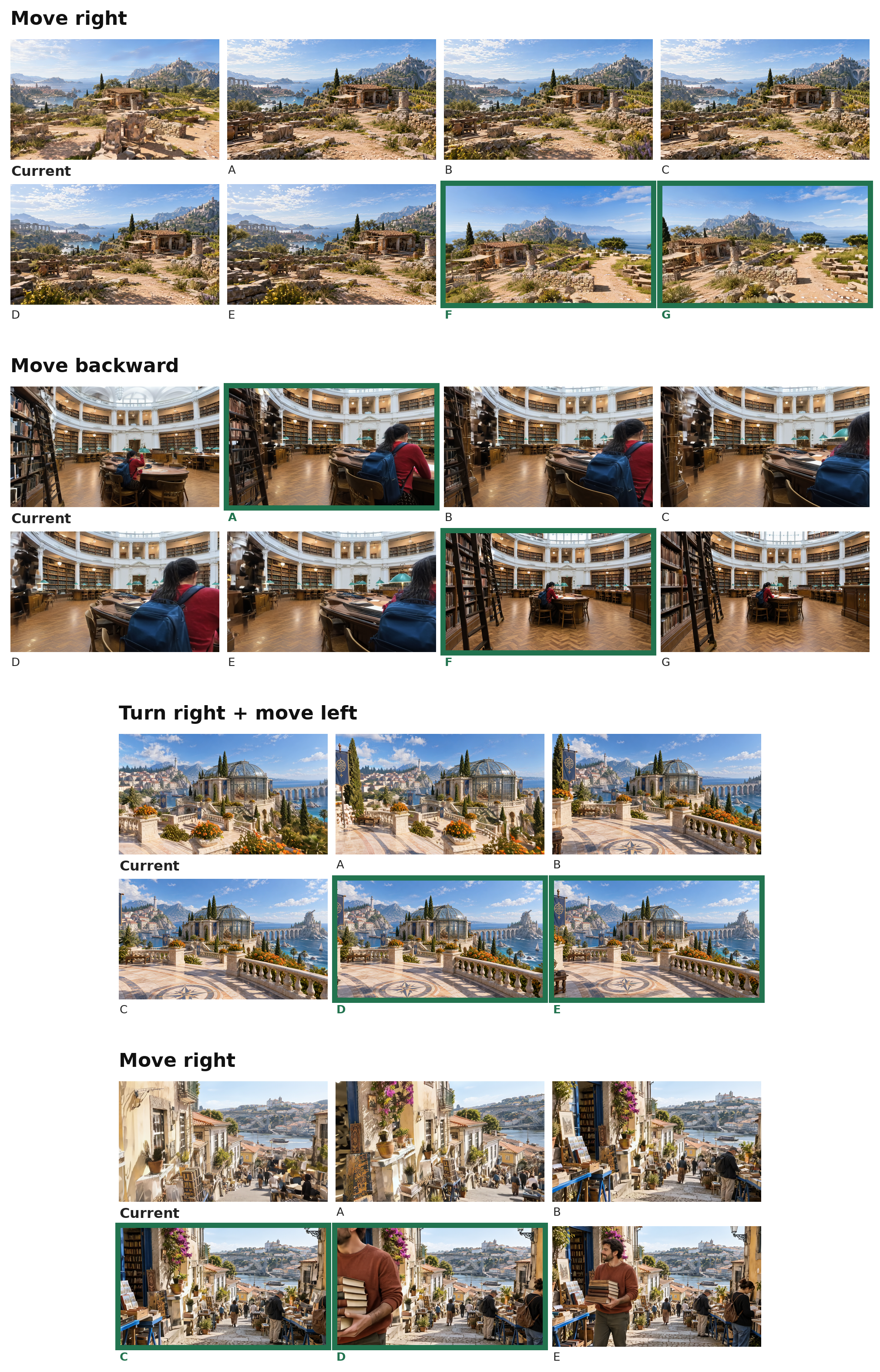}
    \caption{Additional action-aware retrieval examples. `Current' denotes the current observation, and A--G denote historical candidates. Green borders mark the selected candidates.}
    \label{fig:hy15_retrieval_runtime_lateral}
\end{figure}

\begin{figure}[p]
    \centering
    \setlength{\abovecaptionskip}{4pt}
    \setlength{\belowcaptionskip}{0pt}
    \includegraphics[width=0.84\textwidth,height=0.23\textheight,keepaspectratio]{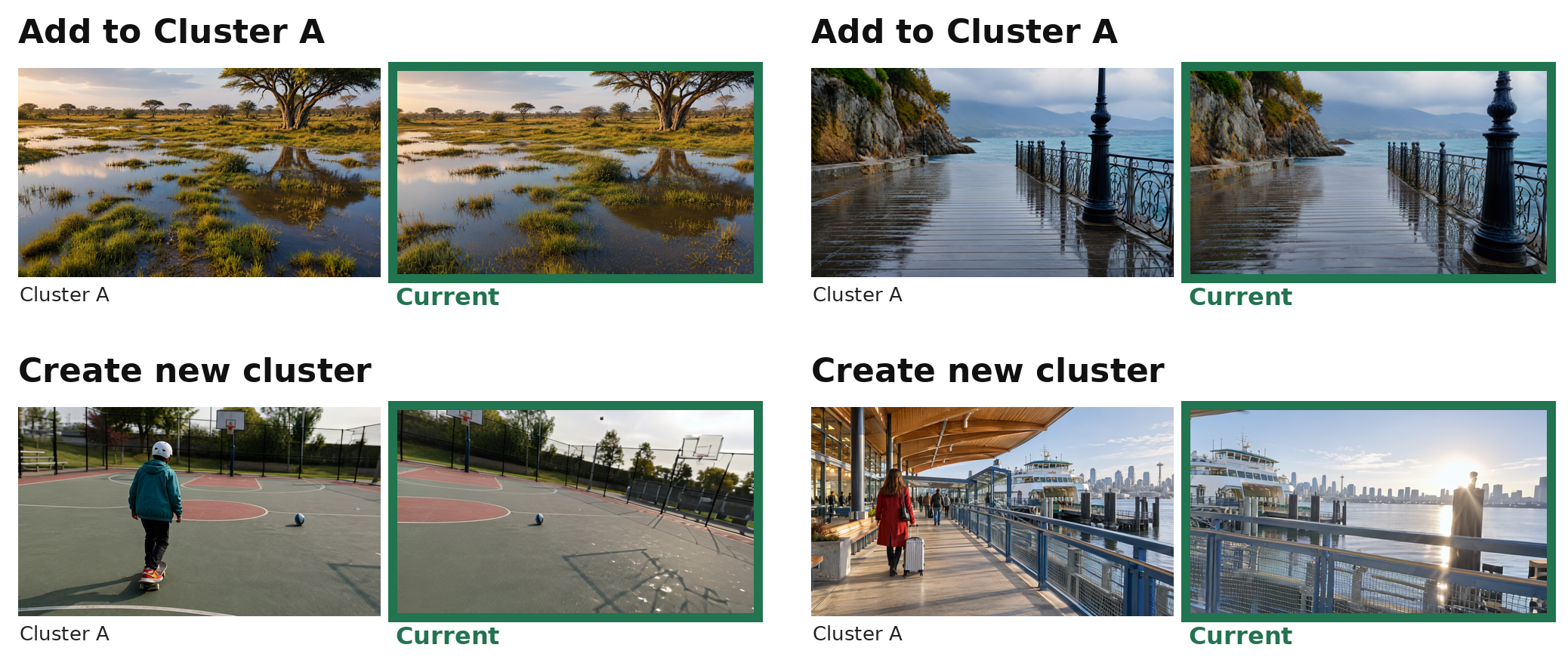}
    \caption{Spatial clustering: add to an existing cluster (top) or create a new cluster (bottom).}
    \label{fig:hy15_spatial_clustering_qual}
    \vspace{5pt}
    \includegraphics[width=0.84\textwidth,height=0.23\textheight,keepaspectratio]{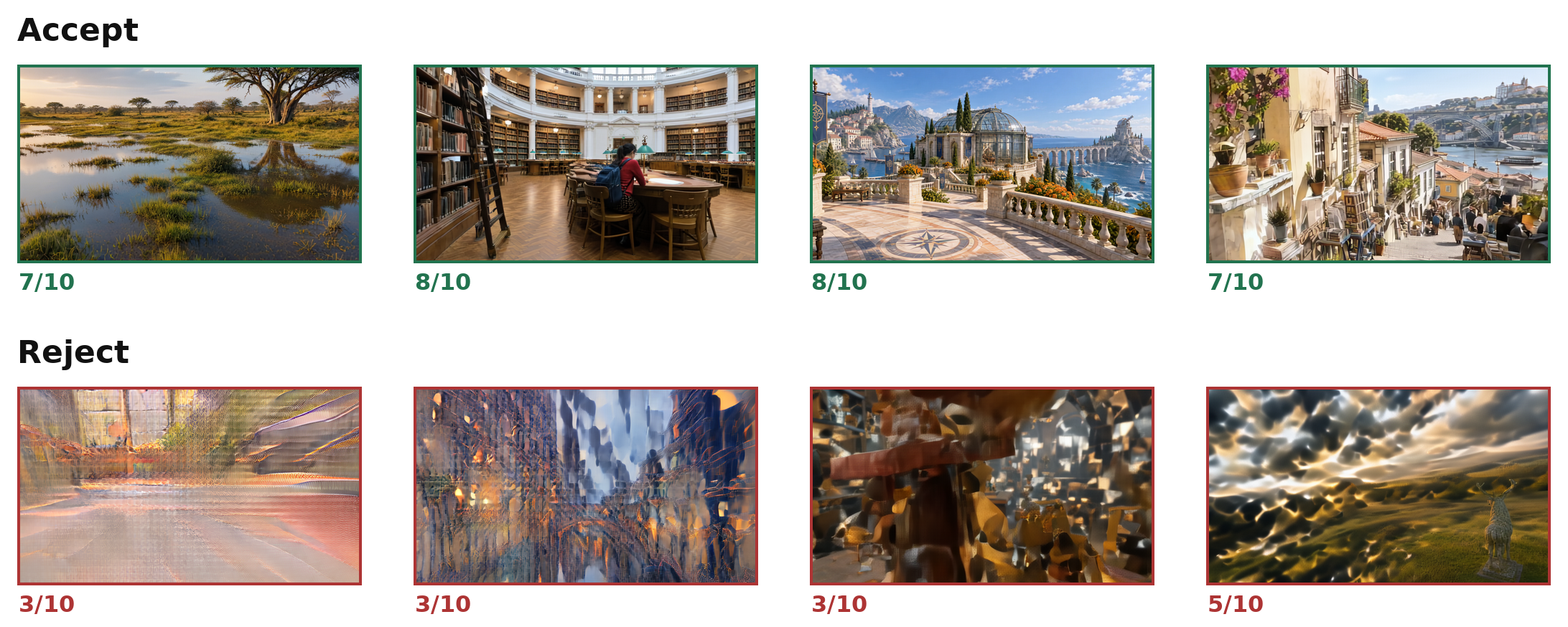}
    \caption{Reliability-aware filtering: accepted chunks (top) and rejected chunks (bottom).}
    \label{fig:hy15_filtering_runtime}
    \vspace{5pt}
    \begin{minipage}[t]{0.49\textwidth}
        \vspace{0pt}\centering
        \includegraphics[width=\linewidth,height=0.38\textheight,keepaspectratio]{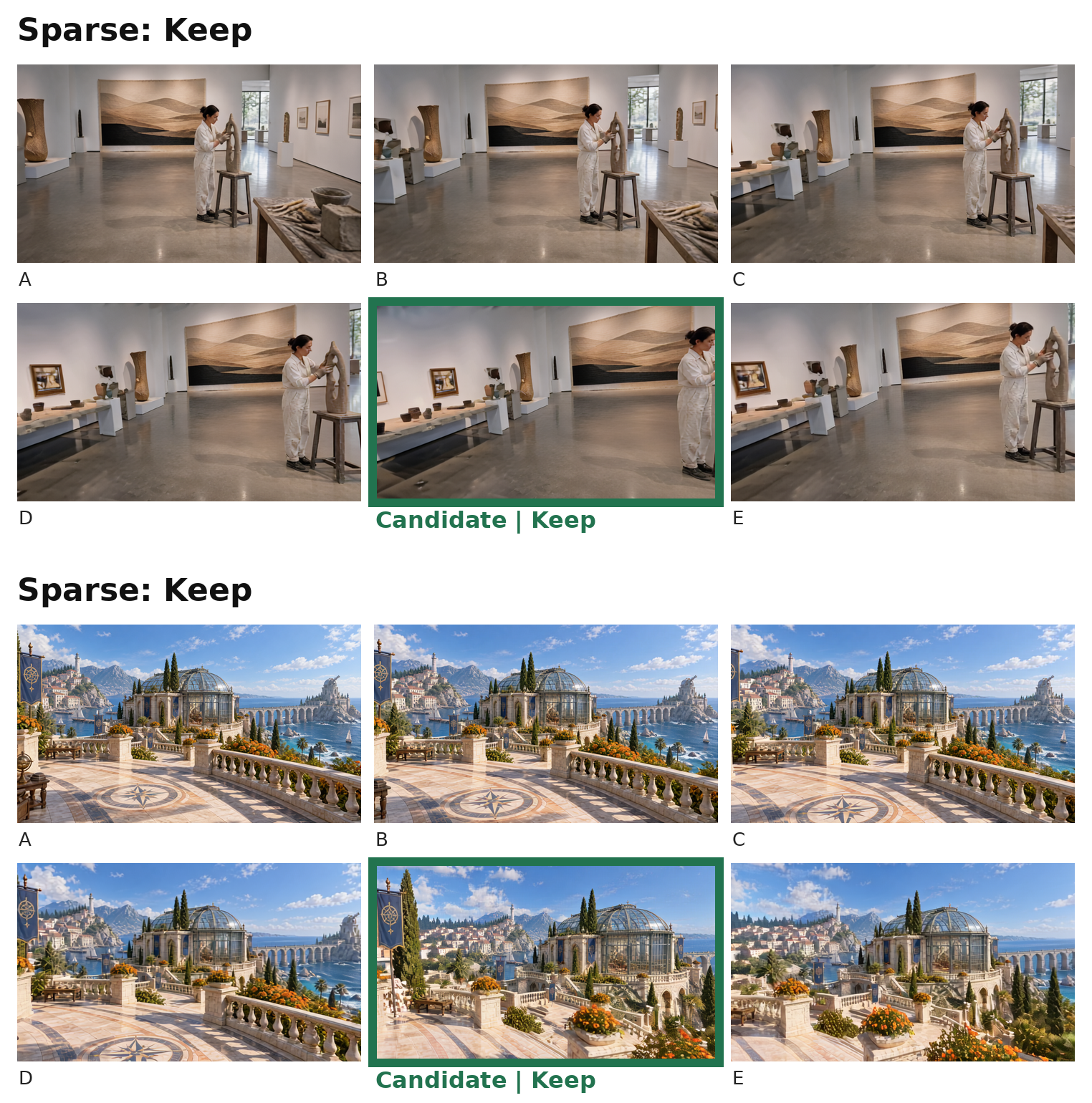}
    \end{minipage}\hfill
    \begin{minipage}[t]{0.49\textwidth}
        \vspace{0pt}\centering
        \includegraphics[width=\linewidth,height=0.38\textheight,keepaspectratio]{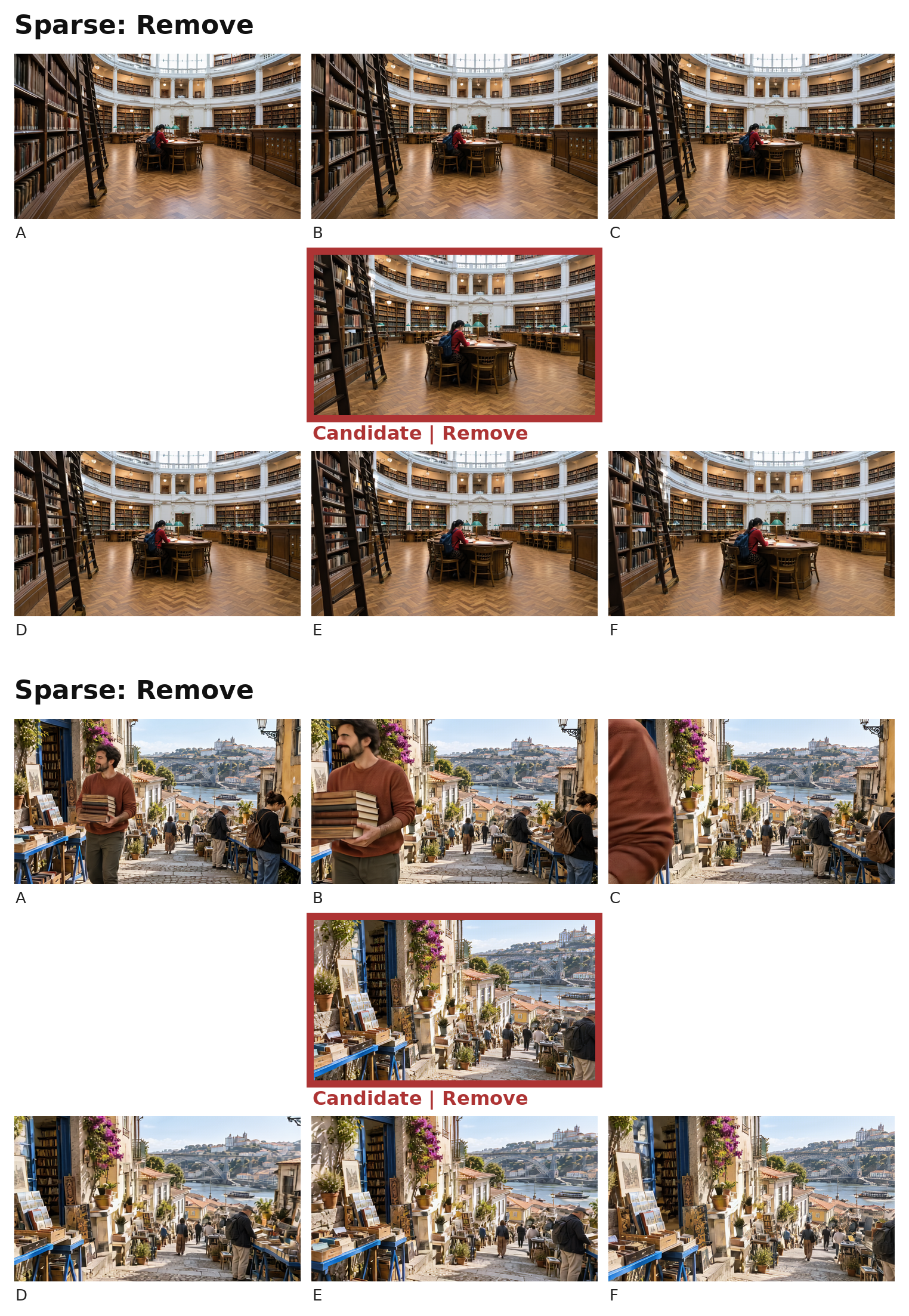}
    \end{minipage}
    \caption{Within-cluster sparsification: KEEP decisions (left) and REMOVE decisions (right).}
    \label{fig:hy15_sparse_runtime}
    \label{fig:hy15_sparse_keep_runtime}
    \label{fig:hy15_sparse_remove_runtime}
\end{figure}

Figures~\ref{fig:hy15_retrieval_runtime} and~\ref{fig:hy15_retrieval_runtime_lateral} visualize action-aware retrieval: conditioned on the current observation and action, the MLLM performs spatial reasoning to select historical candidates that provide appropriate spatial context. Geometry-only retrieval may select irrelevant or incorrect candidates (e.g., A or B) because it does not understand the spatial relationships between observations. Figure~\ref{fig:hy15_spatial_clustering_qual} illustrates assigning a new chunk to an existing cluster or creating a new one; Figure~\ref{fig:hy15_filtering_runtime} shows how chunks are accepted or rejected based on visual reliability; and Figure~\ref{fig:hy15_sparse_runtime} illustrates retaining informative chunks and removing redundant ones through within-cluster sparsification.

\clearpage

\clearpage

\clearpage

\section{Additional Experimental Details}
\label{app:experimental_details}


We implement the four spatial-memory operations on HY1.5 and Wan2.2 as follows.

\paragraph{Spatial clustering.}
We extract the middle frame from the newly generated chunk and from each candidate cluster prototype. The MLLM determines whether they depict the same spatial region. If a match exists, the new chunk is assigned to the corresponding cluster; otherwise, a new cluster is created.

\paragraph{Within-cluster sparsification.}
When the number of chunks in a cluster reaches a predefined threshold, we sample two frames from each chunk. The MLLM compares their spatial content to identify redundancy. An intermediate chunk is removed if it provides no additional spatial structure, object states, or occlusion information relative to its neighboring memories.

\paragraph{Action-aware retrieval.}
We first shortlist the seven historical chunks whose camera positions are closest to the current camera position. The MLLM then considers the spatial location depicted in recent observations and the current action to select historical chunks that provide relevant spatial information for the current generation. Both backbones retain the two most recent chunks as context. No additional chunk is retrieved if none of the candidates is suitable.

\paragraph{Reliability-aware filtering.}
The MLLM scores each newly generated chunk on a scale of 1--10 for visual reliability, with higher scores indicating greater reliability. Chunks scoring below 6 are discarded rather than added to the memory bank, reducing the accumulation of generation errors through subsequent memory reuse.

\section{GPT-5.6-sol Evaluation}
\label{app:vlm_full}

\subsection{Evaluation Protocol}

\paragraph{Input construction.}
For each generated video, we sample one representative frame from every generated chunk and preserve their temporal order to form a long-horizon evaluation sequence. The corresponding action sequence is provided when judging camera-motion adherence. This chunk-level sampling retains the evolution of the generated scene while keeping the evaluation input tractable.

\paragraph{Blind multi-agent scoring.}
We use three independent GPT-5.6-sol evaluator agents. Method identities are hidden, and the candidate videos are presented in randomized order. All evaluators receive the same instruction and independently score the complete sequence. Their predictions are aggregated only after the three evaluations are completed, reducing the influence of ordering and individual evaluator variation.

\paragraph{Evaluation dimensions.}
The protocol assesses adjacent transition continuity, long-range identity preservation, cumulative geometric consistency, overall scene consistency, camera-motion adherence, structural integrity, artifact cleanliness, detail clarity, and overall visual quality. These dimensions jointly capture the consistency, stability, continuity, and visual reliability of long-horizon generation.

\paragraph{Reporting.}
The main paper reports Camera-Constrained Scene Consistency (the Camera-Gated Scene column below) and Overall Visual Quality. Tables~\ref{tab:full_vlm_hy15_results} and~\ref{tab:full_vlm_results} report all nine score dimensions for HY1.5 and Wan2.2. Bold marks the highest score in each column.

\begin{table}[H]
\centering
\caption{Complete GPT-5.6-sol evaluation on HY1.5. All scores are on a 0--100 scale; higher is better.}
\label{tab:full_vlm_hy15_results}
\scriptsize
\setlength{\tabcolsep}{2pt}
\renewcommand{\arraystretch}{1.15}
\resizebox{\textwidth}{!}{%
\begin{tabular}{lrrrrrrrrr}
\hline
Method & \shortstack{Adjacent\\transition} & \shortstack{Long-range\\identity} & \shortstack{Cumulative\\geometry} & \shortstack{Overall scene\\consistency} & \shortstack{Camera-gated\\scene} & \shortstack{Structural\\integrity} & \shortstack{Artifact\\cleanliness} & \shortstack{Detail\\clarity} & \shortstack{Overall visual\\quality} \\
\hline
Base (FoV) & 63.5462 & 51.8905 & 45.5081 & 53.2088 & 51.6053 & 55.9933 & 48.0538 & 53.7048 & 53.1538 \\
FramePack & 56.9633 & 42.8667 & 35.3834 & 44.5193 & 43.2800 & 49.8634 & 40.5100 & 47.8134 & 46.6474 \\
Deep Forcing & 44.2133 & 39.6633 & 28.2400 & 37.9464 & 37.6760 & 37.5733 & 25.1633 & 36.4701 & 33.6297 \\
MoC & 61.4733 & 49.6100 & 42.3133 & 50.7506 & 48.9196 & 54.4600 & 46.8767 & 53.3933 & 51.9717 \\
VMem & 66.0933 & 55.7800 & 48.6767 & 56.5863 & 54.4727 & 59.2067 & 51.3800 & 57.4433 & 56.5060 \\
MemFlow & 56.7033 & 41.1300 & 34.8067 & 43.4457 & 43.3160 & 48.9834 & 39.9500 & 45.8667 & 45.6500 \\
SMI (ours) & \textbf{75.3666} & \textbf{68.4500} & \textbf{62.2167} & \textbf{68.6197} & \textbf{67.0200} & \textbf{68.3000} & \textbf{60.6034} & \textbf{65.1833} & \textbf{65.3677} \\
\hline
\end{tabular}%
}
\end{table}

\begin{table}[H]
\centering
\caption{Complete GPT-5.6-sol evaluation on Wan2.2. All scores are on a 0--100 scale; higher is better.}
\label{tab:full_vlm_results}
\scriptsize
\setlength{\tabcolsep}{2pt}
\renewcommand{\arraystretch}{1.15}
\resizebox{\textwidth}{!}{%
\begin{tabular}{lrrrrrrrrr}
\hline
Method & \shortstack{Adjacent\\transition} & \shortstack{Long-range\\identity} & \shortstack{Cumulative\\geometry} & \shortstack{Overall scene\\consistency} & \shortstack{Camera-gated\\scene} & \shortstack{Structural\\integrity} & \shortstack{Artifact\\cleanliness} & \shortstack{Detail\\clarity} & \shortstack{Overall visual\\quality} \\
\hline
Base (FoV) & 47.8335 & 31.0978 & 29.1101 & 34.8049 & 34.6524 & 39.6402 & 34.8039 & 37.8779 & 37.8363 \\
FramePack & 51.6066 & 30.6701 & 28.8400 & 35.4547 & 34.7984 & 39.7833 & 34.9867 & 38.0200 & 37.9917 \\
Deep Forcing & \textbf{61.6470} & 44.8030 & 43.2600 & \textbf{48.6450} & 46.8560 & 50.8130 & \textbf{49.6300} & 52.1630 & 50.7300 \\
MoC & 59.8633 & \textbf{45.9233} & 43.0867 & 48.2164 & 45.5013 & 48.9634 & 47.7200 & \textbf{52.3930} & 49.2763 \\
VMem & 58.9533 & 41.0967 & 37.2434 & 44.6153 & 43.8800 & 45.6400 & 43.5534 & 47.3770 & 45.3613 \\
MemFlow & 54.3633 & 36.1333 & 31.5034 & 39.5520 & 38.8487 & 40.2234 & 39.1734 & 42.1770 & 40.2990 \\
SMI (ours) & 59.2771 & 42.8278 & \textbf{43.5601} & 47.1430 & \textbf{46.9309} & \textbf{52.8471} & 48.9702 & 51.8770 & \textbf{51.4909} \\
\hline
\end{tabular}%
}
\end{table}

\section{User Study}
\label{app:user_study}

We compare Base, VMem, FramePack, and SMI on 15 scenes, with videos lasting 45--60 seconds under matched initial observations and action sequences. Twenty participants evaluate scene consistency and generation stability. As shown in Table~\ref{tab:user_study_preferences}, SMI leads all compared methods in both dimensions, receiving best-choice preference shares of 52.50\% for scene consistency and 57.83\% for generation stability.

\begin{table}[H]
\centering
\small
\renewcommand{\arraystretch}{1.18}
\setlength{\tabcolsep}{6pt}
\begin{tabular*}{\linewidth}{@{\extracolsep{\fill}}lcccc@{}}
\noalign{\hrule height 0.8pt}
 & Base & VMem & FramePack & SMI (ours) \\
\hline
Scene consistency (\%) & 11.00 & 27.00 & 7.50 & \textbf{52.50} \\
Generation stability (\%) & 11.33 & 27.83 & 2.00 & \textbf{57.83} \\
\noalign{\hrule height 0.8pt}
\end{tabular*}
\caption{Best-choice preference shares (\%) on 15 scenes. Bold indicates the highest share in each row.}
\label{tab:user_study_preferences}
\end{table}

\section{Baseline Implementation Details}
\label{app:baseline_details}
\paragraph{Evaluation settings.}
We compare Base, FramePack, Deep Forcing, Mixture of Contexts (MoC), VMem, MemFlow, and SMI on HY1.5 and Wan2.2, keeping the backbone weights unchanged. Within each backbone, all methods share the same initial observations, action sequences, random seeds, context budget, and generation settings. We retain the two most recent chunks as local context and retrieve at most two additional historical chunks for HY1.5 and three for Wan2.2.


\textbf{Base.} The original backbone retains the local context and retrieves earlier chunks according to field-of-view overlap, up to the backbone-specific cap. It does not use an auxiliary memory manager.

\textbf{FramePack.} We adapt FramePack's temporal hierarchy to the chunk interface. One initial anchor is retained, and older observations are selected by temporal distance using the $4\times$, $2\times$, and $1\times$ tiers. The selected observations follow the native context path; no FramePack-specific encoder or weight is introduced.

\textbf{Mixture of Contexts (MoC).} We adapt the MoC routing idea as a global chunk router at inference. Each chunk contributes a compact block-0 attention descriptor. At each step, the router ranks remote chunks by query-key similarity and combines the highest-scoring chunks with the two local chunks, subject to the backbone-specific cap.

\textbf{Deep Forcing.} Deep Forcing uses a token-level cache with a 10-frame deep-sink prefix and a 4-frame recent query window. When the cache is full, non-sink historical tokens are retained according to attention scores; temporal RoPE positions are compensated after reorganization. We use the strict token-cache implementation and do not use the separate random-selection or compression-free variants.

\textbf{VMem.} Our VMem implementation builds a surfel map from geometry estimated by LoGeR~\citep{zhang2026loger} and scores historical chunks by their contribution to the target-view rendering. It combines the two local chunks with at most two additional chunks on HY1.5 or three on Wan2.2, and then re-encodes the selected chunks through the native context path.

\textbf{MemFlow.} We adapt MemFlow's semantic-retrieval component to the native chunk interface. Every second chunk contributes its first decoded RGB frame as a candidate, and the last decoded frame of the preceding chunk is used as the query. We rank candidates by CLIP~\citep{radford2021clip} image-image cosine similarity, exclude the two local chunks, and retrieve at most two remote chunks on HY1.5 or three on Wan2.2.

\section{Operation-Oriented Data Construction}
\label{app:data_construction}

\paragraph{Collection and manual annotation.}
We first collect approximately one thousand videos, each longer than one minute, from actual world-model inference and manually remove low-quality videos. We then manually annotate 354 samples in total across the four operations, providing reliable supervision examples for the subsequent annotation stage.

\paragraph{Teacher annotation and quality control.}
Using the manually annotated samples as few-shot demonstrations, GPT-5.6-sol and Gemini-3.1 Pro annotate the remaining data. Claude Fable 5 then examines the annotations produced by the two teacher MLLMs and flags anomalous outputs. Finally, we further review and clean the flagged samples to obtain the complete training dataset.

\paragraph{Operation-specific supervision.}
The resulting samples are organized according to the four atomic operations. Reliability-aware filtering predicts the visual reliability of a newly generated chunk. Spatial clustering determines whether a new chunk matches an existing spatial prototype. Within-cluster sparsification determines whether a candidate chunk is redundant given the remaining memories in the same cluster. Action-aware retrieval selects historical chunks that can support generation under the current context and planned action. Each sample pairs its operation-specific multimodal input $\mathcal{I}_n$ with a structured target output $Y_n$.

\paragraph{Dataset composition.}
Because HY1.5 and Wan2.2 have different generation characteristics, we construct a separate supervision dataset for each backbone. Table~\ref{tab:operation_data_composition} reports the number of training samples for every operation. The HY1.5 training set contains 63,911 samples in total, including 24,623 samples for action-aware retrieval (Coverage), accounting for approximately 38.53\% of the training set.

\begin{table}[h]
\centering
\caption{Composition of the operation-oriented training datasets.}
\label{tab:operation_data_composition}
\small
\setlength{\tabcolsep}{8pt}
\begin{tabular}{lrr}
\hline
Operation & HY1.5 & Wan2.2 \\
\hline
Spatial clustering & 12,652 & 12,538 \\
Within-cluster sparsification & 7,156 & 7,135 \\
Action-aware retrieval & 24,623 & 6,852 \\
Reliability-aware filtering & 19,480 & 4,411 \\
\hline
Total & 63,911 & 30,936 \\
\hline
\end{tabular}
\end{table}

\subsection{Operation-Wise Label Composition}
\label{app:operation_label_composition}

We further report the supervision-label composition of each atomic operation. Table~\ref{tab:operation_label_composition} summarizes the label distributions for spatial clustering, within-cluster sparsification, and reliability-aware filtering. All statistics below are computed over the effective training occurrences used for optimization.

\paragraph{Spatial clustering.}
Each training sample compares a newly generated chunk with one candidate spatial prototype. A \textit{belongs} label indicates that the two observations describe the same local spatial region, whereas \textit{does not belong} indicates that they should not be assigned to the same cluster. HY1.5 contains 9,041 belongs samples and 3,611 does-not-belong samples; Wan2.2 contains 8,972 and 3,566 samples, respectively.

\paragraph{Within-cluster sparsification.}
Each sample evaluates a candidate chunk relative to the remaining memories in the same cluster. A \textit{keep} label indicates that the chunk retains useful spatial evidence, while \textit{remove} indicates that it is redundant. The HY1.5 dataset contains 2,045 keep samples and 5,111 remove samples, whereas Wan2.2 contains 2,051 and 5,084 samples, respectively.

\paragraph{Reliability-aware filtering.}
The visual-quality task assigns a reliability score from 1 to 10 to a newly generated chunk. Following the filtering rule used by SMI, chunks with scores no greater than 6 are labeled as filtered, while chunks with scores greater than 6 are retained. Under this binary grouping, 24.55\% of HY1.5 samples and 38.79\% of Wan2.2 samples are filtered. The remaining 75.45\% and 61.21\%, respectively, are retained. These statistics characterize the supervision distribution rather than the runtime filtering rate.

\begin{table}[t]
\centering
\caption{Label composition of the spatial clustering, within-cluster sparsification, and reliability-aware filtering training data.}
\label{tab:operation_label_composition}
\small
\setlength{\tabcolsep}{6pt}
\begin{tabular}{llrr}
\hline
Operation & Label & HY1.5 & Wan2.2 \\
\hline
Spatial clustering & Belongs & 9,041 (71.46\%) & 8,972 (71.56\%) \\
Spatial clustering & Does not belong & 3,611 (28.54\%) & 3,566 (28.44\%) \\
Within-cluster sparsification & Keep & 2,045 (28.58\%) & 2,051 (28.75\%) \\
Within-cluster sparsification & Remove & 5,111 (71.42\%) & 5,084 (71.25\%) \\
Reliability-aware filtering & Filtered ($\leq 6$) & 4,783 (24.55\%) & 1,711 (38.79\%) \\
Reliability-aware filtering & Retained ($>6$) & 14,697 (75.45\%) & 2,700 (61.21\%) \\
\hline
\end{tabular}
\end{table}

\paragraph{Action-aware retrieval.}
Coverage supervision conditions memory selection on the planned camera motion. Table~\ref{tab:coverage_motion_composition} reports the motion distribution over 24,623 HY1.5 and 6,852 Wan2.2 training occurrences. The percentages therefore describe the effective exposure seen during training, including motion balancing. For HY1.5, 15 and 9 legacy records with non-standard motion strings are merged into the \texttt{tilt\_down} and \texttt{tilt\_up} categories, respectively. 

\begin{table}[t]
\centering
\caption{Motion composition of the action-aware retrieval training data.}
\label{tab:coverage_motion_composition}
\scriptsize
\setlength{\tabcolsep}{7pt}
\begin{tabular}{lrr}
\hline
Motion & HY1.5 & Wan2.2 \\
\hline
\texttt{mostly\_static} & 0.75\% & 2.01\% \\
\texttt{move\_backward} & 9.36\% & 7.41\% \\
\texttt{move\_forward} & 12.50\% & 7.68\% \\
\texttt{move\_left} & 10.20\% & 7.59\% \\
\texttt{move\_right} & 9.81\% & 7.41\% \\
\texttt{tilt\_down} & 2.70\% & 4.25\% \\
\texttt{tilt\_up} & 2.47\% & 4.44\% \\
\texttt{tilt\_down+move\_backward} & 1.87\% & 2.85\% \\
\texttt{tilt\_down+move\_forward} & 1.91\% & 3.12\% \\
\texttt{tilt\_down+move\_left} & 0.75\% & 1.69\% \\
\texttt{tilt\_down+move\_right} & 0.75\% & 1.59\% \\
\texttt{tilt\_up+move\_backward} & 0.75\% & 1.72\% \\
\texttt{tilt\_up+move\_forward} & 1.87\% & 2.47\% \\
\texttt{tilt\_up+move\_left} & 0.75\% & 1.71\% \\
\texttt{tilt\_up+move\_right} & 1.87\% & 1.88\% \\
\texttt{turn\_left} & 8.72\% & 5.65\% \\
\texttt{turn\_right} & 8.57\% & 6.67\% \\
\texttt{turn\_left+move\_backward} & 1.87\% & 3.30\% \\
\texttt{turn\_left+move\_forward} & 1.23\% & 3.65\% \\
\texttt{turn\_left+move\_left} & 1.87\% & 3.30\% \\
\texttt{turn\_left+move\_right} & 1.62\% & 4.28\% \\
\texttt{turn\_left+tilt\_down} & 0.19\% & 0.03\% \\
\texttt{turn\_left+tilt\_up} & 0.19\% & 0.07\% \\
\texttt{turn\_right+move\_backward} & 1.87\% & 3.50\% \\
\texttt{turn\_right+move\_forward} & 1.22\% & 3.59\% \\
\texttt{turn\_right+move\_left} & 1.44\% & 4.67\% \\
\texttt{turn\_right+move\_right} & 1.87\% & 3.44\% \\
\texttt{turn\_right+tilt\_down} & 0.19\% & 0.03\% \\
Unclassified motion metadata & 10.86\% & 0.00\% \\
\hline
\end{tabular}
\end{table}

\section{Prompts for the Four Atomic Operations}
\label{app:operation_prompts}

We use four operation-specific prompts for reliability-aware filtering, spatial clustering, within-cluster sparsification, and action-aware retrieval. The templates below preserve the decision rules used at runtime; implementation paths, hashes, and dynamically populated image and geometry fields are omitted.

\begin{promptbox}{Reliability-Aware Filtering}
\textbf{Role and input.} Evaluate one frame from a newly generated chunk; when multiple frames are provided, judge the overall visual reliability of the chunk.\par
\textbf{Task.} Assign a visual-quality score from 1 to 10 indicating whether the generated chunk is reliable enough to serve as a future memory reference. Judge only visible corruption, blur, structural artifacts, unstable geometry, distorted layout, broken regions, and overall reference reliability. An unusual scene, object, or viewpoint should not be penalized unless it contains a visible quality failure.\par
\textbf{Scale.} Scores 1--2 denote an unusable result; 3--4 denote obvious artifacts or instability; 5 denotes a marginal reference; 6--7 denote a mostly reliable reference with minor defects; 8--9 denote a clean and stable reference; and 10 denotes an excellent reference without meaningful defects.\par
\textbf{Output.} Return strict JSON containing only the integer field \texttt{visual\_quality\_score}.
\end{promptbox}

\begin{promptbox}{Action-Aware Retrieval}
\textbf{Input.} The final frame of the immediately preceding chunk, the planned camera motion, and candidate historical final frames with their labels, chunk identifiers, and precomputed visual-quality scores.\par
\textbf{Task.} Judge which candidates can support the expected next view through viewpoint coverage, visible-boundary continuity, spatial support, wider context, or recent dynamic-state continuity. Evaluate every candidate independently.\par
\textbf{Motion grouping.} Assign every candidate exactly once to \texttt{supports}, \texttt{neutral}, or \texttt{opposes}. Motion alignment and coverage usefulness are separate judgments; an opposing candidate may remain useful when it provides strong spatial support, although an equally useful supporting or neutral candidate receives higher priority. Forward or nearly static motion favors high-overlap stable references; backward or widening motion favors complementary broad coverage; and lateral or turning motion favors support on the corresponding side of the expected next view.\par
\textbf{Acceptance and selection.} Put every useful reference in \texttt{acceptable}. Then choose at most two labels from this set for \texttt{selected}, prioritizing the strongest next-view support and complementary coverage while avoiding near duplicates. When no candidate is useful, both lists are empty. Visual-quality scores are reliability hints rather than the sole selection criterion.\par
\textbf{Output.} Return strict JSON containing only the three motion groups, \texttt{acceptable}, and \texttt{selected}.
\end{promptbox}

\begin{promptbox}{Spatial Clustering}
\textbf{Input.} The middle frame of the new chunk, the middle frame of one existing class prototype, and \texttt{prototype\_fov\_overlap} in $[0,1]$ as a geometric hint.\par
\textbf{Task.} Decide whether the new chunk and the prototype belong to the same local viewpoint range. This is only the clustering decision; no sparsification or memory-admission decision is made here. Semantic similarity alone is insufficient: the two views must share concrete visible area, compatible local layout, and viewpoint continuity.\par
\textbf{Labels and rules.} Assign \texttt{view\_overlap\_score} from 1 to 6, where 1 indicates no compatible local overlap and 6 indicates very strong spatial alignment. Set \texttt{match=false} when the score is at most 2 and \texttt{match=true} when it is at least 3. The FoV overlap is used as a hint rather than the sole criterion.\par
\textbf{Output.} Return strict JSON containing only \texttt{match} and \texttt{view\_overlap\_score}.
\end{promptbox}

\begin{promptbox}{Within-Cluster Sparsification}
\textbf{Input.} One middle frame from every remaining chunk in the same cluster, the middle frame of the candidate chunk, and \texttt{coverage\_by\_remaining\_chunks} in $[0,1]$ as a geometric hint.\par
\textbf{Task.} Remove the candidate only when the remaining chunks already cover its useful spatial information and it contributes no useful information beyond them. Otherwise, keep it.\par
\textbf{Labels.} Predict \texttt{spatial\_coverage} from \{none, weak, medium, strong, full\} and \texttt{unique\_useful\_information} from \{none, small, clear\}. Unique information may include an unrepresented viewpoint, boundary, layout, transition, occlusion, dynamic state, or local spatial relation that is useful for future consistency.\par
\textbf{Decision rules.} Output \texttt{remove} only when coverage is strong or full and unique useful information is none or small. Keep the candidate when coverage is none, weak, or medium, or whenever unique useful information is clear. The numeric coverage hint supports the decision but never determines it alone.\par
\textbf{Output.} Return strict JSON containing only \texttt{action}, \texttt{spatial\_coverage}, and \texttt{unique\_useful\_information}.
\end{promptbox}

\clearpage
\section{Additional Qualitative Comparisons}
\label{app:additional_comparisons}

\begin{figure}[!ht]
    \centering
    \makebox[0pt][l]{\raisebox{\dimexpr0.385\textheight-7pt\relax}[0pt][0pt]{\hspace{4pt}\small\bfseries (a)}}%
    \includegraphics[width=\textwidth,height=0.385\textheight,keepaspectratio]{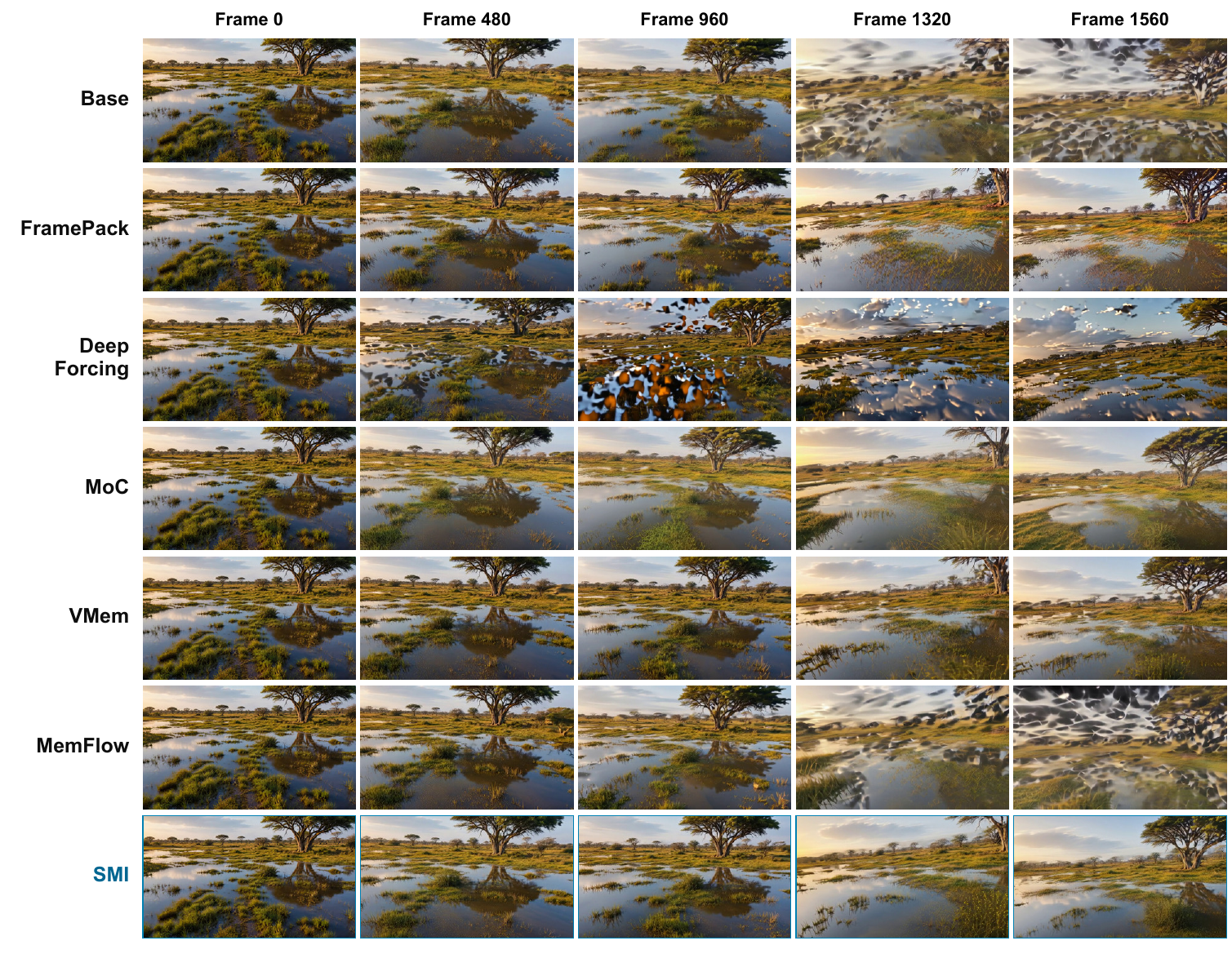}\par
    \vspace{5pt}
    \makebox[0pt][l]{\raisebox{\dimexpr0.385\textheight-7pt\relax}[0pt][0pt]{\hspace{4pt}\small\bfseries (b)}}%
    \includegraphics[width=\textwidth,height=0.385\textheight,keepaspectratio]{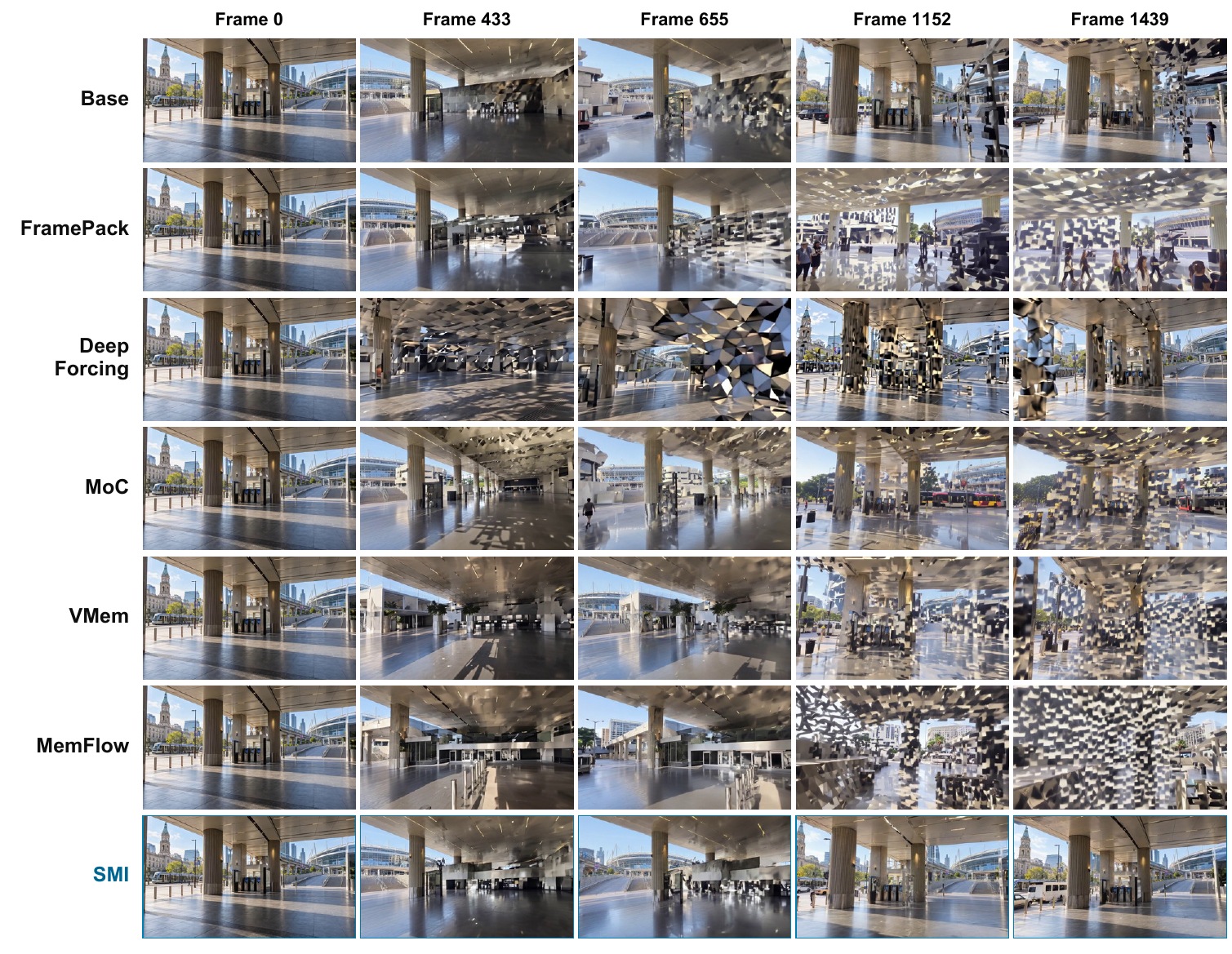}
    \caption{Additional qualitative results on HY1.5.}
    \label{fig:hy15_seven_methods_stability}
\end{figure}

\begin{figure}[p]
    \centering
    {\small\bfseries (a)\par}
    \vspace{2pt}
    \includegraphics[width=\textwidth,height=0.44\textheight,keepaspectratio]{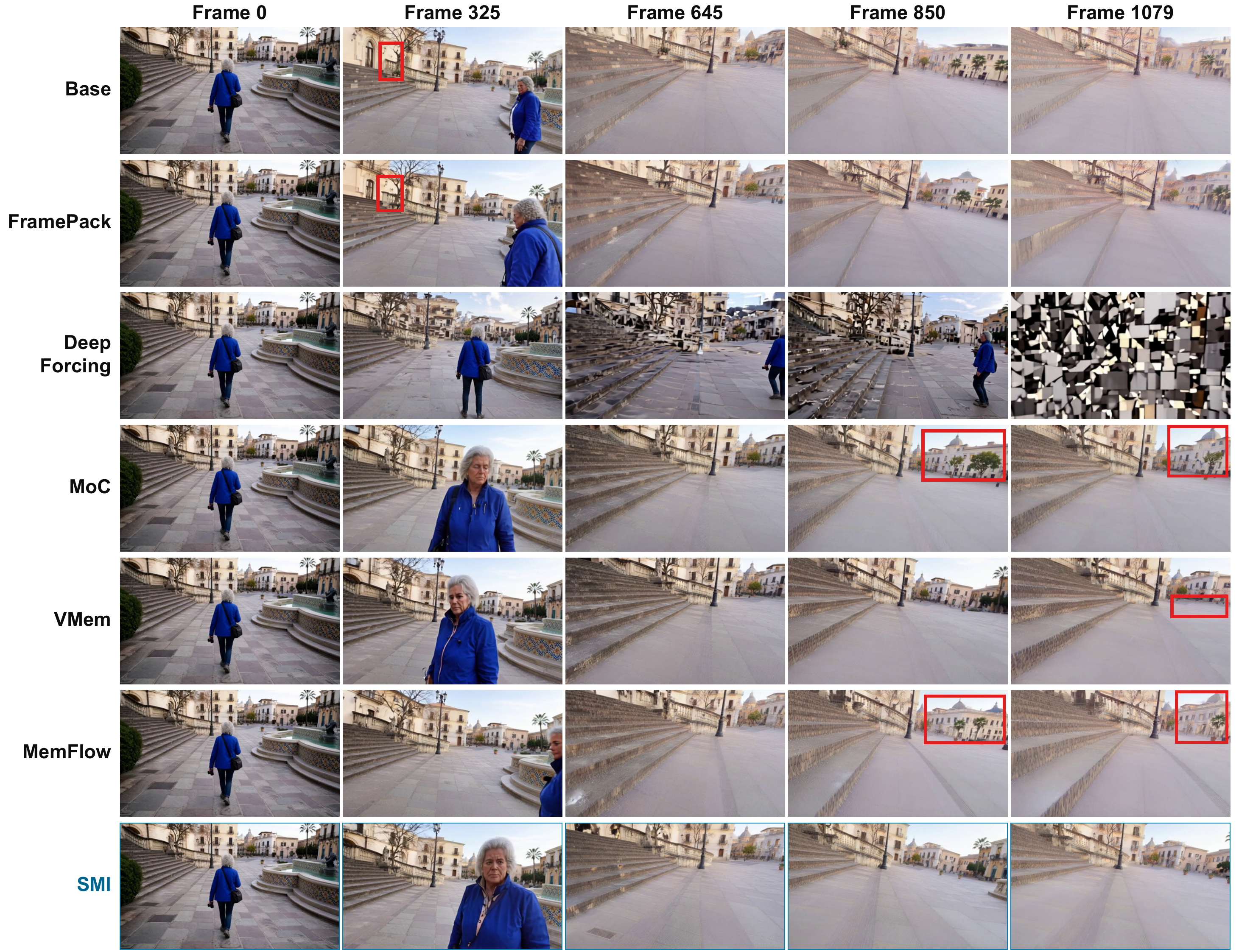}\par
    \vspace{5pt}
    {\small\bfseries (b)\par}
    \vspace{2pt}
    \includegraphics[width=\textwidth,height=0.44\textheight,keepaspectratio]{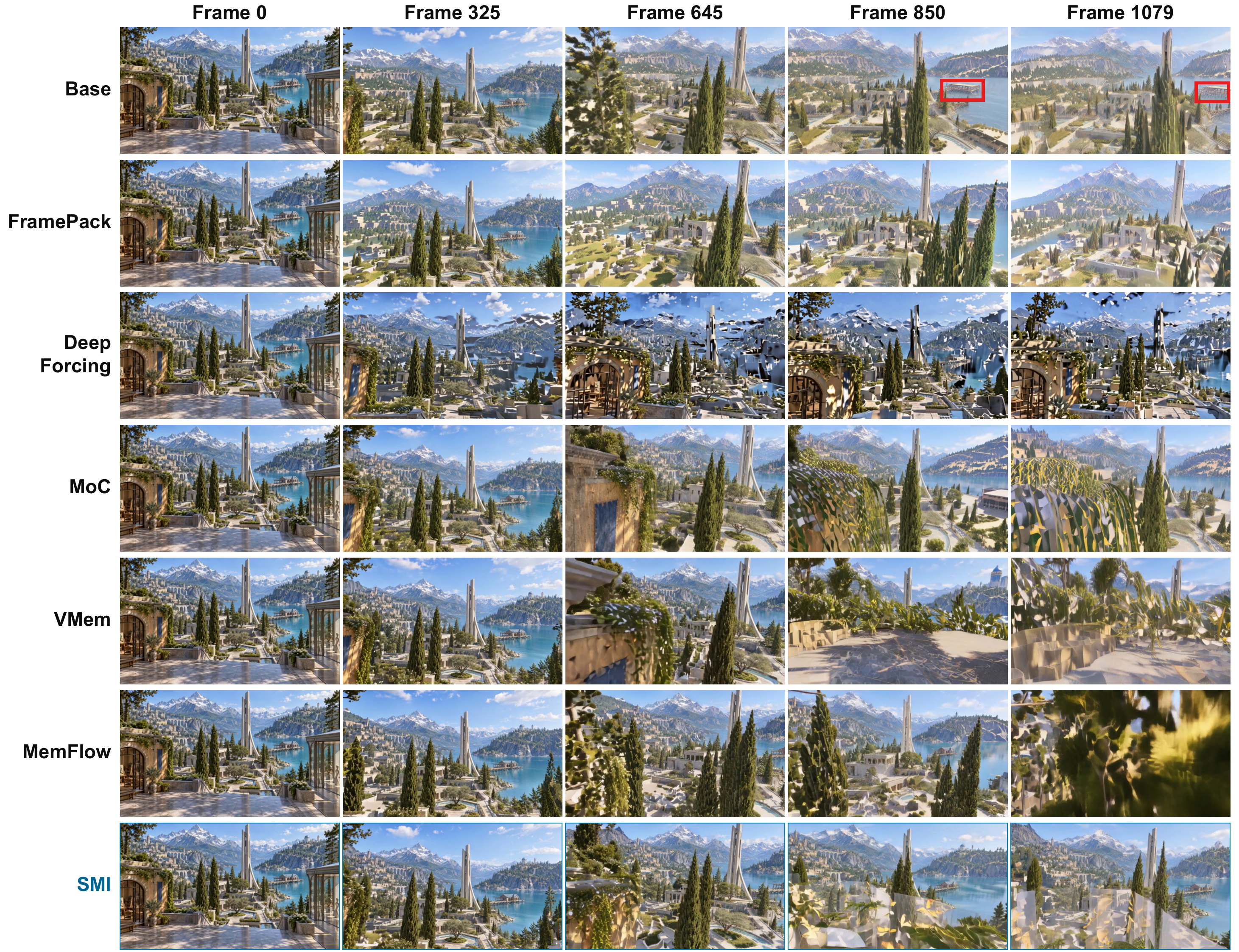}
    \caption{More qualitative results on HY1.5. Red boxes mark spatial inconsistencies.}
    \label{fig:hy15_plaza_lakeside}
\end{figure}

\begin{figure}[p]
    \centering
    {\small\bfseries (a)\par}
    \vspace{2pt}
    \includegraphics[width=\textwidth,height=0.44\textheight,keepaspectratio]{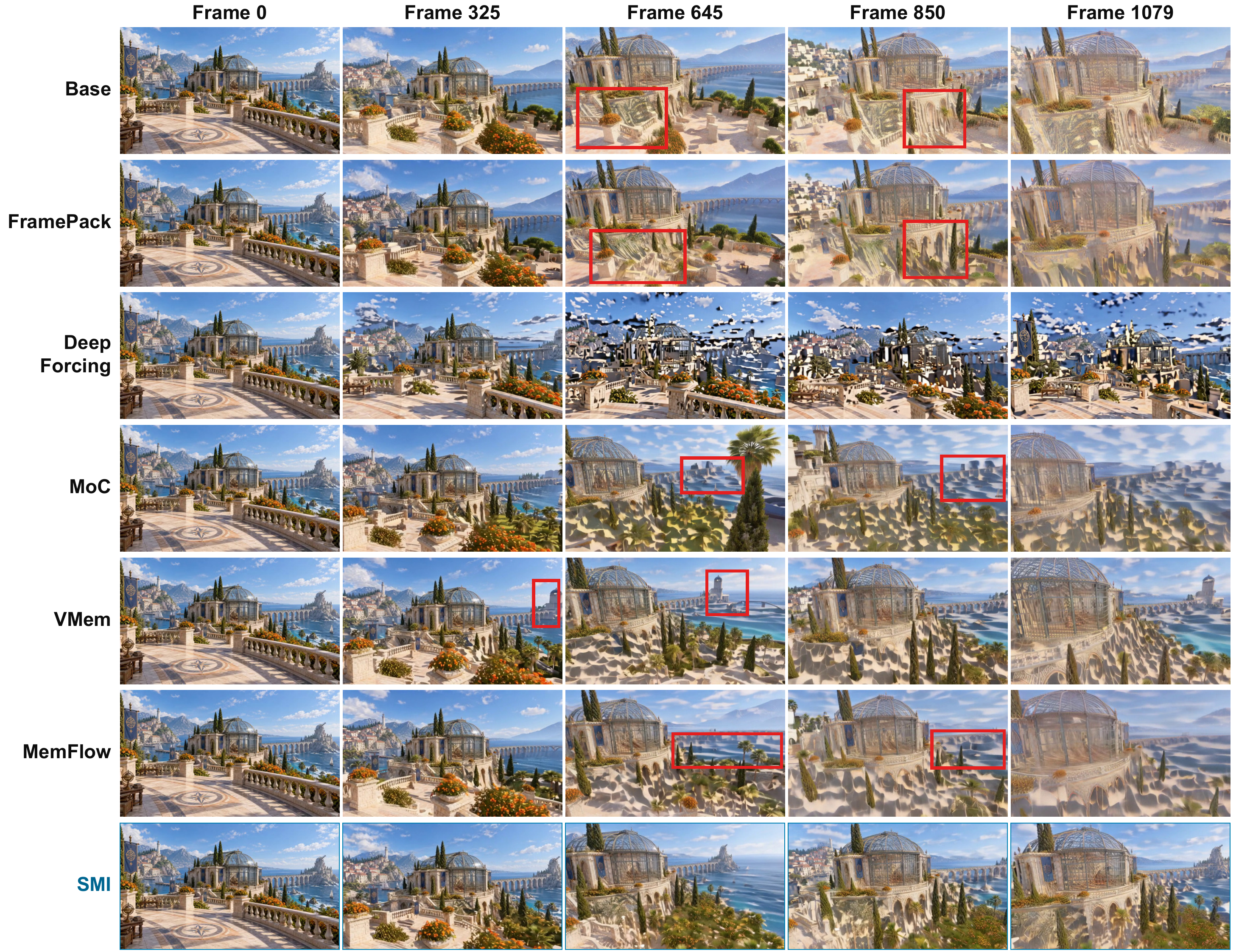}\par
    \vspace{5pt}
    {\small\bfseries (b)\par}
    \vspace{2pt}
    \includegraphics[width=\textwidth,height=0.44\textheight,keepaspectratio]{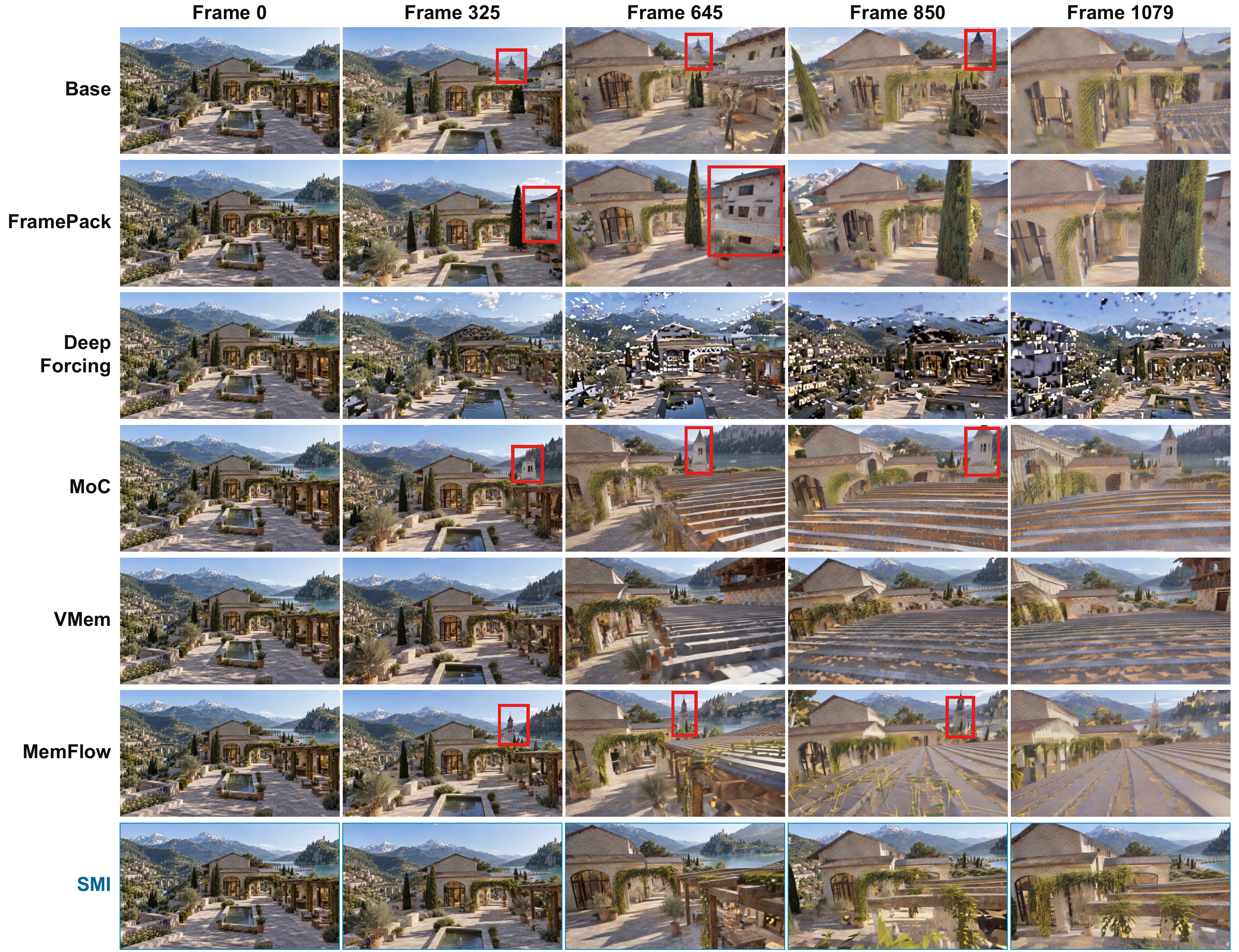}
    \caption{More qualitative results on HY1.5. Red boxes mark spatial inconsistencies.}
    \label{fig:hy15_bridge_courtyard}
\end{figure}

\begin{figure}[p]
    \centering
    {\small\bfseries (a)\par}
    \vspace{2pt}
    \includegraphics[width=\textwidth,height=0.44\textheight,keepaspectratio]{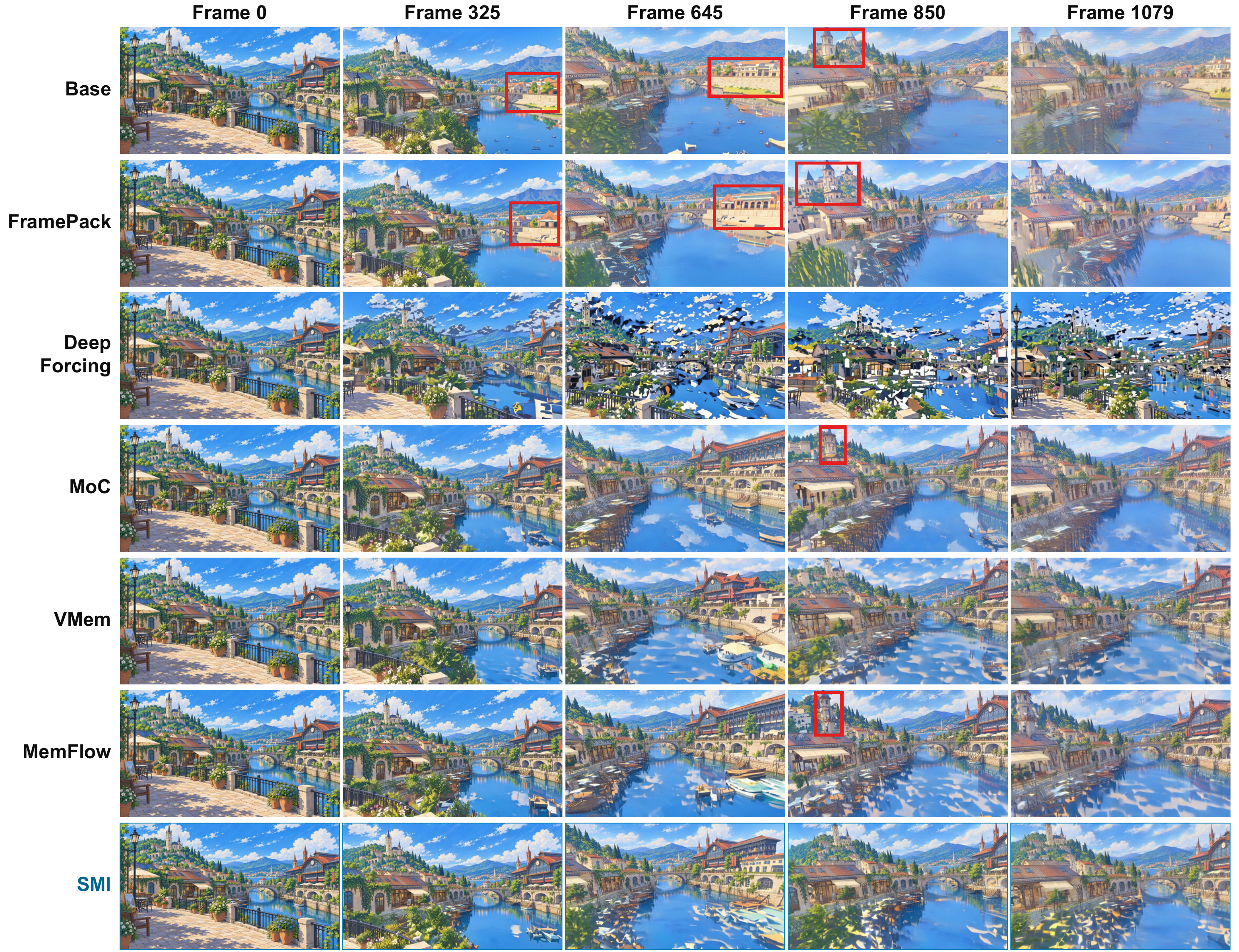}\par
    \vspace{5pt}
    {\small\bfseries (b)\par}
    \vspace{2pt}
    \includegraphics[width=\textwidth,height=0.44\textheight,keepaspectratio]{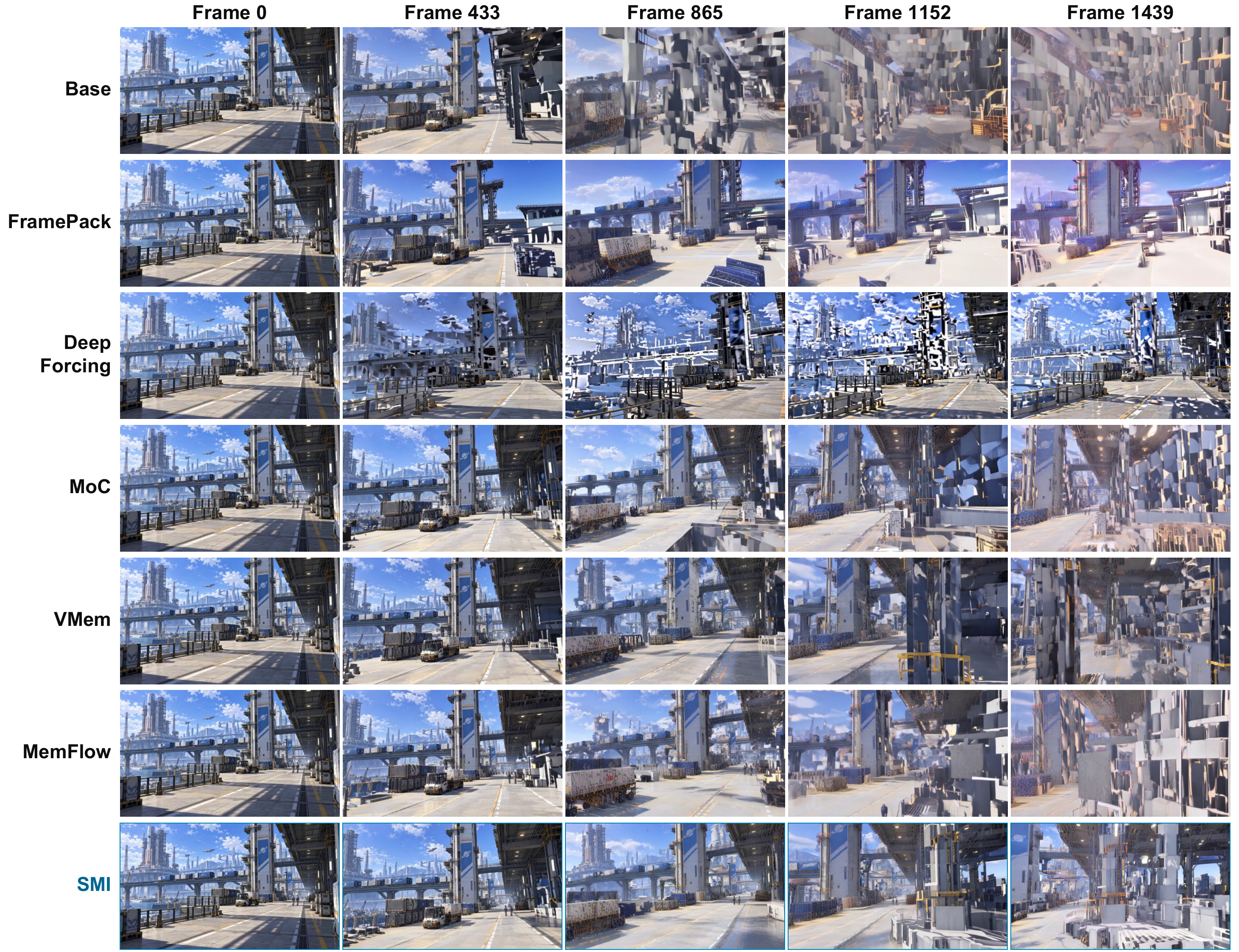}
    \caption{More qualitative results on HY1.5. Red boxes mark spatial inconsistencies.}
    \label{fig:hy15_canal_port}
\end{figure}

\begin{figure}[!ht]
    \centering
    {\small\bfseries (a)\par}
    \vspace{2pt}
    \includegraphics[width=\textwidth,height=0.41\textheight,keepaspectratio]{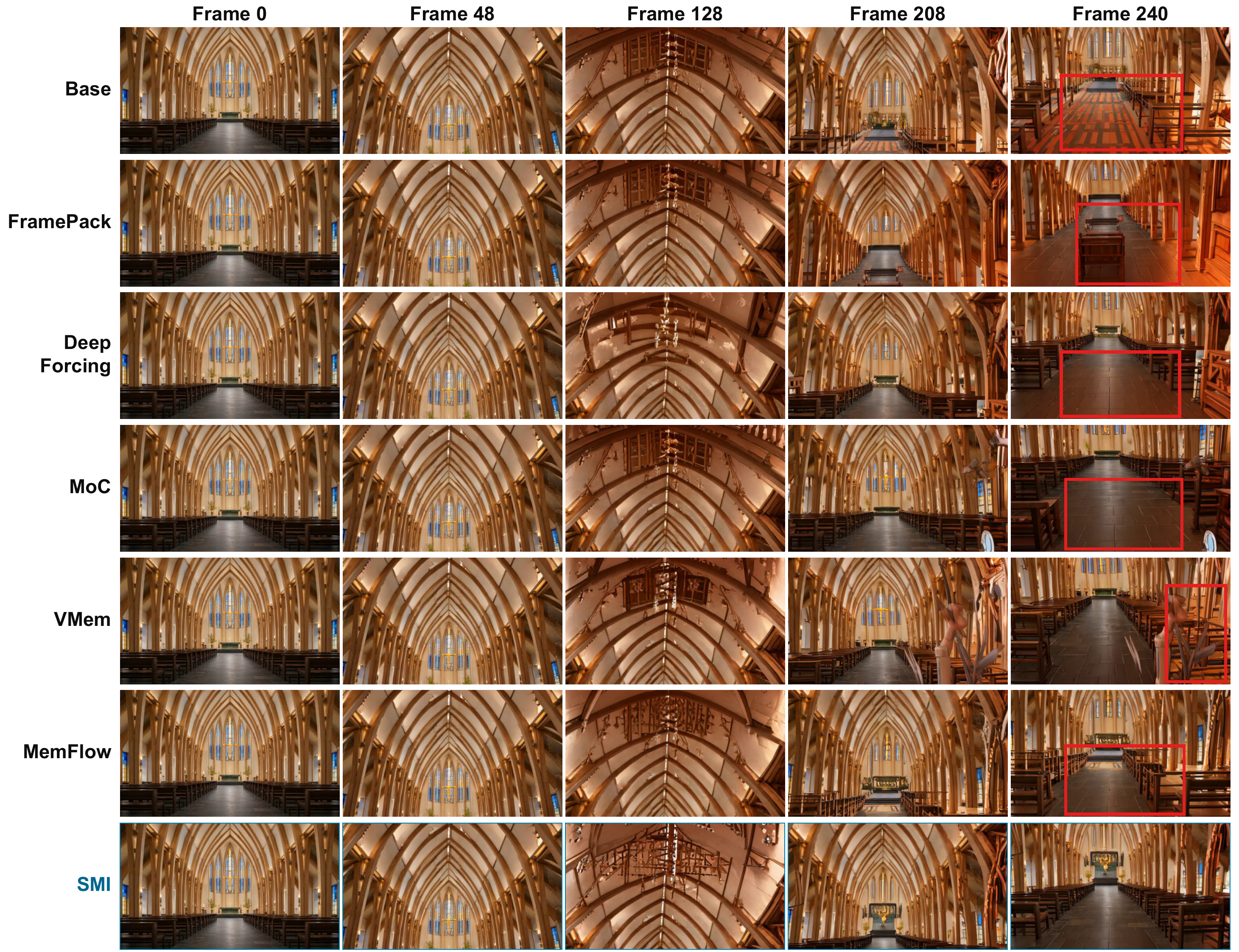}\par
    \vspace{5pt}
    {\small\bfseries (b)\par}
    \vspace{2pt}
    \includegraphics[width=\textwidth,height=0.41\textheight,keepaspectratio]{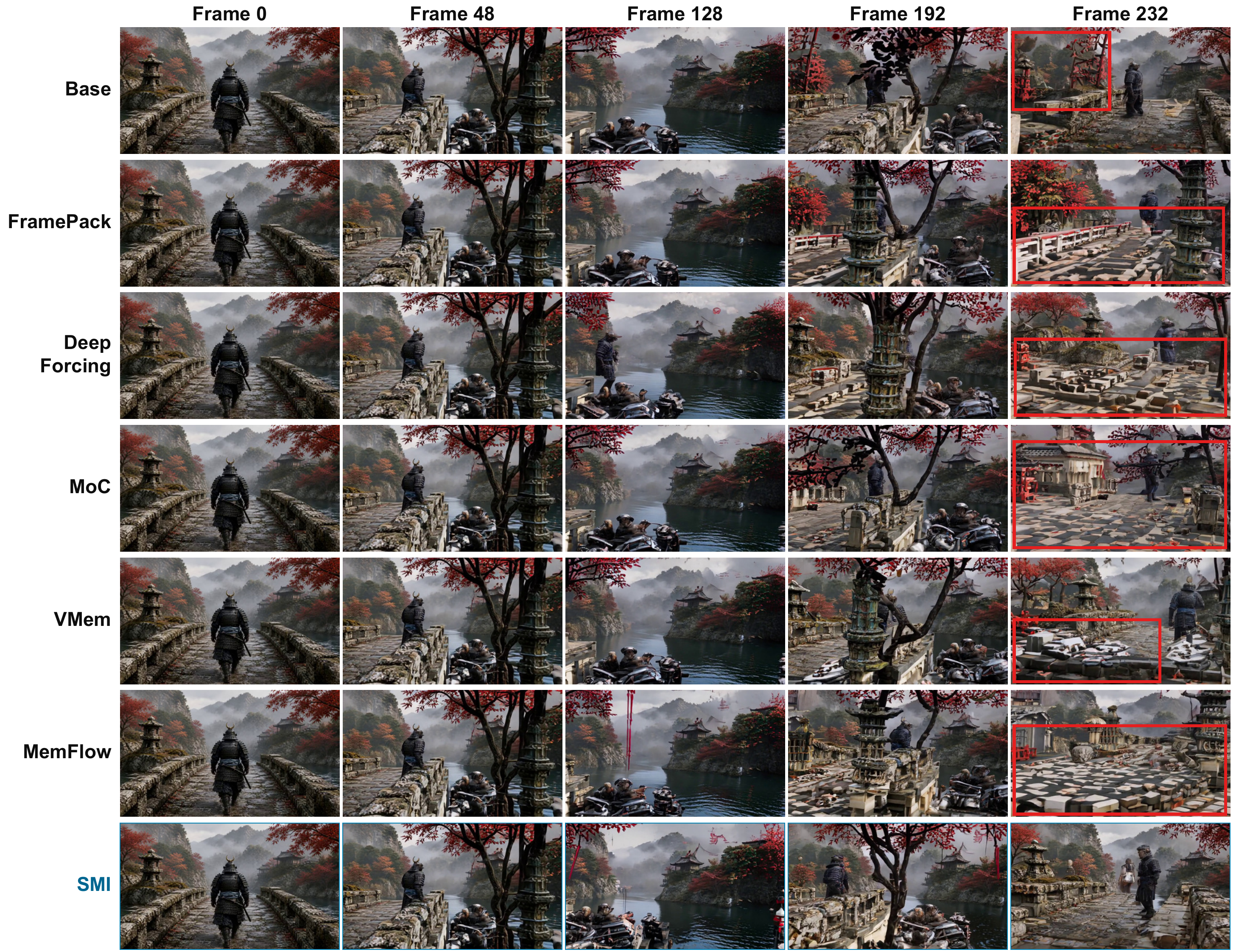}
    \caption{Additional qualitative results on Wan2.2.}
    \label{fig:wan22_methods_consistency_01}
    \label{fig:wan22_methods_consistency_03}
\end{figure}

\begin{figure}[!ht]
    \centering
    {\small\bfseries (a)\par}
    \vspace{2pt}
    \includegraphics[width=\textwidth,height=0.44\textheight,keepaspectratio]{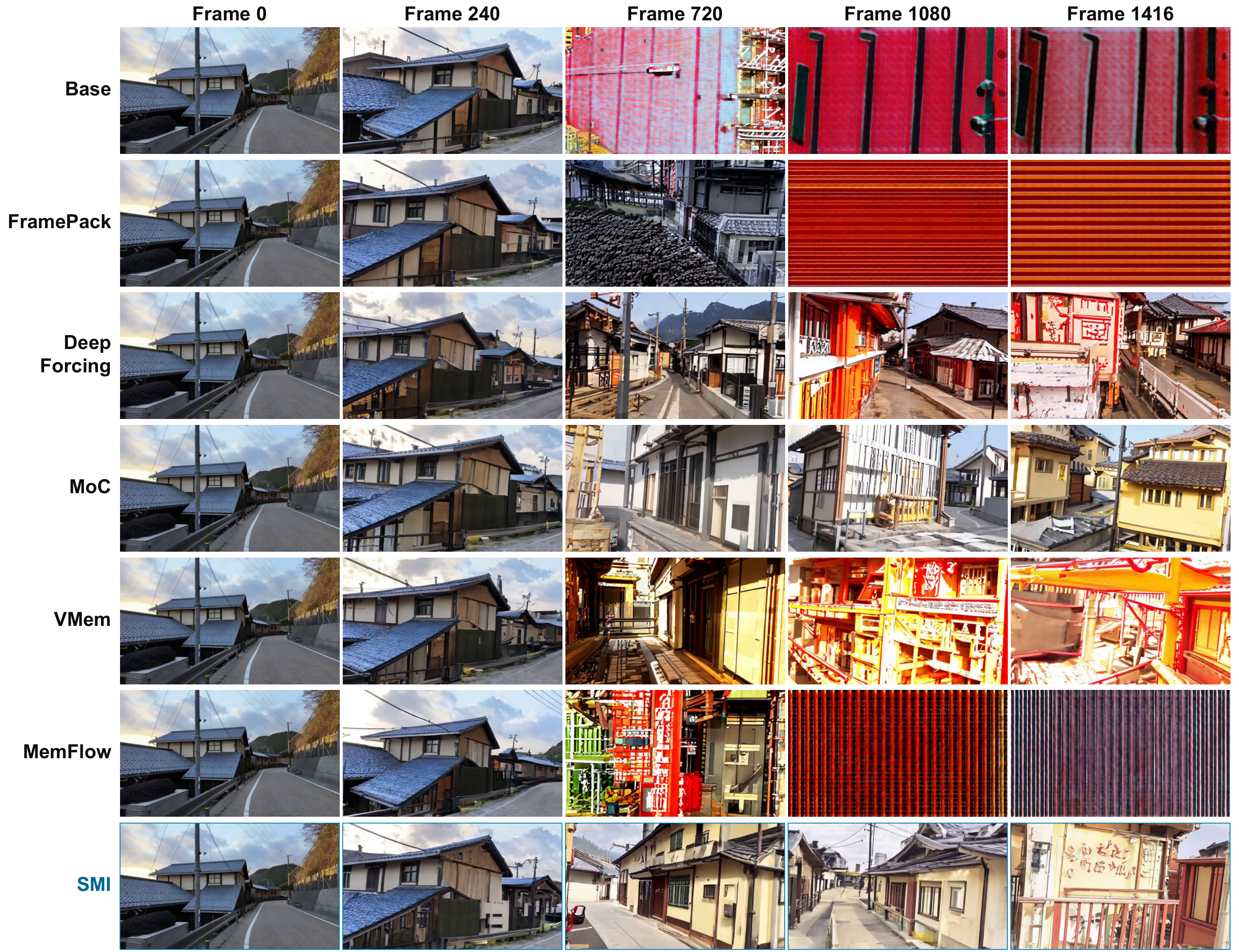}\par
    \vspace{5pt}
    {\small\bfseries (b)\par}
    \vspace{2pt}
    \includegraphics[width=\textwidth,height=0.44\textheight,keepaspectratio]{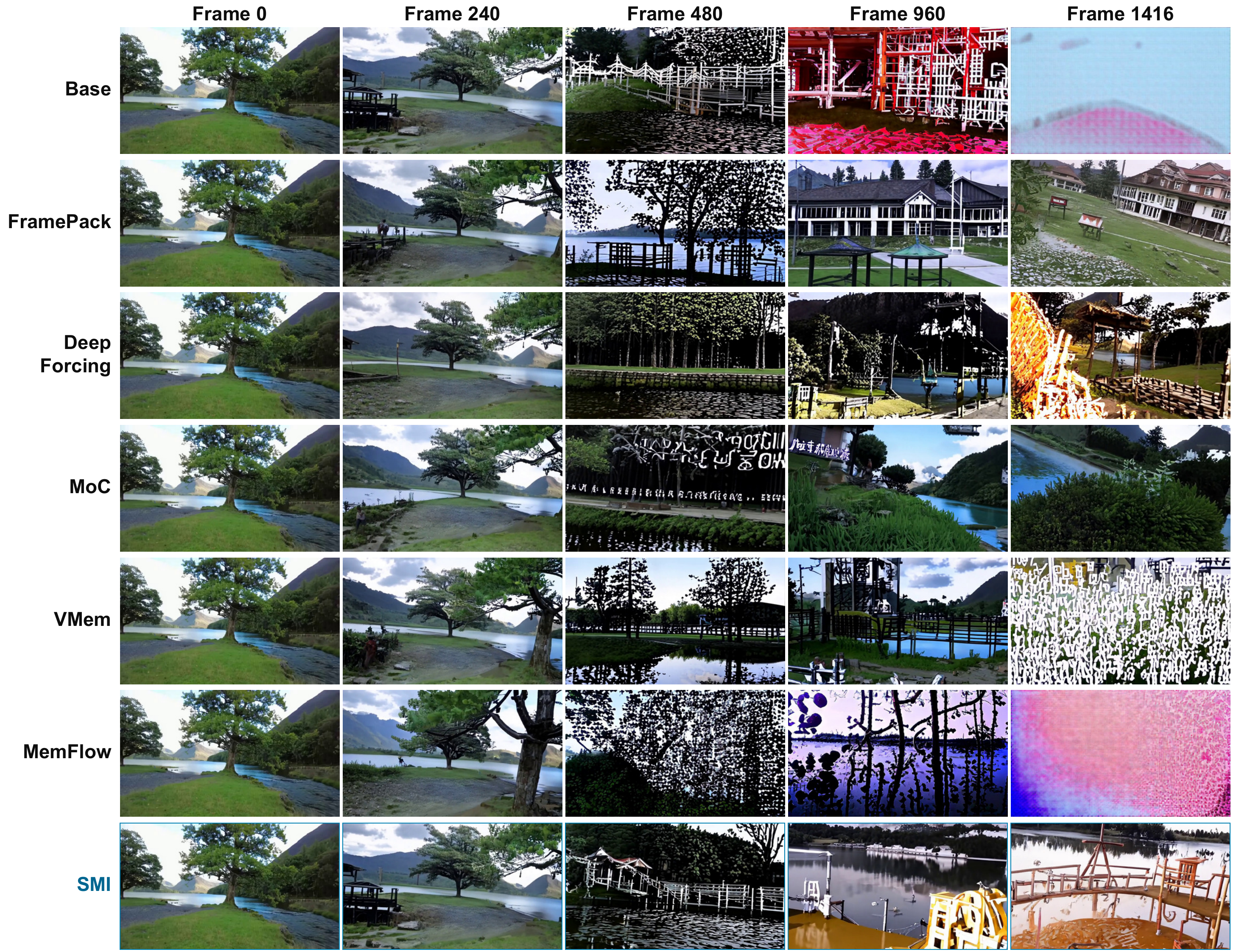}
    \caption{More qualitative results on Wan2.2.}
    \label{fig:wan22_methods_stability_01}
    \label{fig:wan22_methods_stability_02}
\end{figure}

We provide additional qualitative comparisons on HY1.5 (Figures~\ref{fig:hy15_seven_methods_stability}--\ref{fig:hy15_canal_port}) and Wan2.2 (Figures~\ref{fig:wan22_methods_consistency_01}--\ref{fig:wan22_methods_stability_01}). SMI achieves comprehensive improvements in generation stability and spatial consistency.

\subsection{HY1.5}
\label{app:hy15_additional_comparisons}

\clearpage

\clearpage

\clearpage

\clearpage

\subsection{Wan2.2}
\label{app:wan22_additional_comparisons}

\label{app:wan22_seven_methods}
All seven methods use the same frame indices within each case. Red boxes mark visible inconsistencies in the final baseline frames.

\clearpage

%
%
%
%

\end{document}